\documentclass[letterpaper,twocolumn,10pt]{article}
\usepackage{usenix} 

\usepackage{graphicx}
\usepackage{amsmath}
\usepackage{amssymb}
\usepackage{amsthm}
\usepackage{colortbl}
\usepackage[vlined,ruled,linesnumbered]{algorithm2e}
\usepackage{caption}
\usepackage{booktabs}
\usepackage{capt-of}
\usepackage{balance}
\usepackage[caption=false,font=footnotesize]{subfig}
\usepackage{multirow}
\usepackage{siunitx}
\usepackage{makecell}
\usepackage{tabularx}
\usepackage{tabularray}
\usepackage{threeparttable}
\usepackage[]{hyperref}
\usepackage[nameinlink,noabbrev]{cleveref}
\usepackage{enumitem}
\usepackage[normalem]{ulem}
\usepackage{marvosym}
\usepackage{bm}
\usepackage[available,functional,reproduced]{usenixbadges}
\definecolor{lightpink}{RGB}{255, 229, 204}
\definecolor{lightblue}{RGB}{204, 230, 255}
\definecolor{lighterpink}{RGB}{255, 242, 230}
\definecolor{lighterblue}{RGB}{230, 242, 255}
\definecolor{lightergreen}{RGB}{230, 252, 235}

\SetCommentSty{commentfont}
\SetKwComment{Comment}{~$\triangleright$~}{}
\SetSideCommentRight
\SetFillComment

\newcommand{\customsize}{\scriptsize}
\newenvironment{smalltabularx}{\customsize \tabularx}{\endtabularx}
\newcolumntype{C}{>{\centering\arraybackslash\hsize=.5\hsize\linewidth=\hsize}X}

\newcommand{\impup}[1]{\textcolor{green!50!black}{↑\,#1\%}}
\newcommand{\impdown}[1]{\textcolor{red!70!black}{↓\,#1\%}}
\newcommand{\impnone}{\textcolor{gray}{0.00\%}}
\newcommand{\stripminus}[1]{\expandafter\the\numexpr-#1}

\newtheorem{proposition}{Proposition}

\newcommand{\proref}[1]{\hyperref[#1]{Proposition~\ref*{#1}}}
\newcommand{\aref}[1]{\hyperref[#1]{Appendix~\ref*{#1}}}

\setlist[itemize]{leftmargin=*,itemsep=-0.1em,topsep=0.2em, partopsep=0.2em}

\newcommand{\noul}[1]{\vspace{1.5pt}\noindent\uline{#1}}
\newcommand{\nobf}[1]{\vspace{1.5pt}\noindent\textbf{#1.}}
\newcommand{\sys}{\textsc{Retrofit}}

\colorlet{tcolor}{black}
\newcommand{\rcolor}[1]{\textcolor{tcolor}{#1}}

\newenvironment{tightdisplay}{%
  \setlength{\abovedisplayskip}{2pt}%
  \setlength{\belowdisplayskip}{2pt}%
  \setlength{\abovedisplayshortskip}{1pt}%
  \setlength{\belowdisplayshortskip}{1pt}%
}{}

\renewenvironment{equation}
  {\begin{tightdisplay}\begin{oldequation}}
  {\end{oldequation}\end{tightdisplay}}

\date{}  
\title{\Large \bf \sys{}: Continual Learning with \rcolor{Controlled} Forgetting for Binary Security  Detection and Analysis} 

\author{
{\rm Yiling He$^{1,\star}$\Letter, Junchi Lei$^{2,\star}$, Hongyu She$^{2}$, }\\
{\rm Shuo Shao$^{2}$, Xinran Zheng$^{1}$, Yiping Liu$^{3}$, Zhan Qin$^{2}$, Lorenzo Cavallaro$^{1}$}\\
$^{1}$ University College London, London, United Kingdom\\
$^{2}$ Zhejiang University, Hangzhou, China\\
$^{3}$ Alibaba Group, Beijing, China\\
}

\begin{document}
\maketitle  

\begin{abstract}
Binary security has increasingly relied on deep learning to reason about malware behavior and program semantics. 
However, the performance often degrades as threat landscapes evolve and code representations shift.
While continual learning (CL) offers a natural solution through sequential updates, most existing approaches rely on data replay or unconstrained updates, limiting their applicability and effectiveness in data-sensitive security environments.
We propose \sys{}, which regulates knowledge retention and adaptation with \rcolor{controlled} forgetting at each update, without requiring historical data.
Our key idea is to consolidate previously trained and newly fine-tuned models, serving as teachers of legacy and emergent knowledge through retrospective-free parameter merging.
\rcolor{Forgetting control} is achieved by 1)~constraining parameter changes to low-rank and sparse subspaces for approximate orthogonality, and 2)~employing a confidence-guided arbitration to dynamically aggregate knowledge from both teachers.

Our evaluation on two representative applications demonstrates that \sys{} consistently mitigates forgetting while maintaining adaptability.
In malware detection under temporal drift, it substantially improves the retention score, from 20.2\% to 38.6\% over CL baselines, and exceeds the oracle upper bound on new data.
In binary summarization across decompilation levels, where analyzing stripped binaries is especially challenging, \sys{} achieves over 2× the BLEU score of transfer learning used in prior work and surpasses all baselines in cross-representation generalization.

\end{abstract}

{
  \renewcommand{\thefootnote}{\fnsymbol{footnote}}
  \setcounter{footnote}{0}
  \footnotetext{$\star$~Co-first authors. \Letter~Corresponding author: heyilinge0@gmail.com}
}


\section{Introduction}

Deep learning models have become integral to modern binary security systems, driving advances in applications such as malware detection~\cite{qian2025lamd} and binary analysis~\cite{lau2024revisiting, su2024source}. 
However, their effectiveness is constantly challenged by the non-stationarity from two dominant sources: \emph{temporal drift}, as malware families and attack techniques evolve over time~\cite{yang2021bodmas}, and \emph{representation shift}, as binaries undergo compilation, optimization, and symbol stripping~\cite{cao2024evaluating}.
Such variability can destabilize learned decision boundaries, resulting in critical alarm failures or even catastrophic system compromises~\cite{ibm2025threatindex}.

Continual learning (CL) offers a natural paradigm for adapting models to this evolving data, aiming to accumulate knowledge across a sequence of domains rather than training a static model on a fixed dataset~\cite{wang2024comprehensive}. In security analytics, this paradigm is already well-recognized for addressing temporal drift in malware detection, where models are periodically updated through sequential retraining~\cite{continuous} or replay-based strategies~\cite{rahman2025madar}. 
We further identify that representation shift, where data differ across abstraction levels (e.g., increasingly stripped binaries), poses an analogous challenge.
While prior work often employs transfer learning~\cite{ying2018transfer} for adaptation between single domains, we argue that CL's cumulative knowledge provides a distinct advantage. 
As illustrated in~\autoref{fig:motivate_binary}, even a basic CL approach enables stronger forward transfer to each subsequent task than independent single-task fine-tuning.

Nevertheless, applying CL in security-critical scenarios is highly constrained.
Strict data-retention and sharing regulations for malware samples and proprietary binaries, together with progressively arriving large-scale data, make dominant strategies such as full retraining or data replay infeasible~\cite{lai2019enabling, raynal2025conflict}.
This setting fundamentally amplifies two coupled challenges to effective model adaptation. 
First, models struggle to preserve prior knowledge without old data, leading to \emph{catastrophic forgetting}~\cite{hayes2020remind}.
We find this issue to be surprisingly severe in malware detection: as validated in~\autoref{sec:eva_cmp_cl}, adapting to new threats causes striking forgetting of older yet still-relevant malware, with the F1-score of all existing methods dropping below~0.8 within three years.
Second, models fail to build a \emph{cumulative understanding} to integrate new knowledge. 
As a result, they lack robust cross-representation generalization: experience gained from simpler data cannot be leveraged to enhance performance on more complex, data-scarce tasks like analyzing stripped binaries~\cite{al2023extending}.

To address these challenges, we observe that both retention loss and generalization failure stem from the same root cause: uncontrolled model updates that overwrite previously learned information. 
We therefore propose \sys{}, a data \uline{Retro}spective-\uline{F}ree CL method that achieves effective \uline{I}nformation \uline{T}ransfer through \rcolor{controlled} forgetting. 

\sys{} replaces inaccessible historical data with the previous model as a proxy for past knowledge, satisfying the data constraints inherent to security applications.
Knowledge accumulation is then formulated as model consolidation, where a newly updated model learns new information and is subsequently integrated with the previous one via parameter-level merging.
To realize \rcolor{controlled} forgetting in this process, \sys{} employs two complementary mechanisms. 
\emph{Structurally}, it confines parameter updates to task-relevant subspaces via low-rank decomposition. 
\emph{Functionally}, it optimizes a sparse attribution mask that adaptively balances knowledge contributions guided by model confidence.

We evaluate \sys{} on two applications, each exemplifying a distinct form of non-stationarity. 
For temporal drift, we consider Android malware detection~\cite{jordaney2017transcend}, a long-standing benchmark for CL where models must adapt to evolving threats without forgetting earlier ones.
For representation shift, we study binary analysis~\cite{al2023extending}, a more recent LLM-driven summarization task in which models are sequentially trained on decompiled Linux binaries of increasing obfuscation levels to assess cross-representation generalization.
The two datasets comprise 259K labeled apps collected across five years and 215K decompiled function–documentation pairs spanning multiple optimization conditions, respectively.

Across both applications, we demonstrate the clear superiority of \sys{} over five representative CL baselines~\cite{mccloskey1989catastrophic, rebuffi2017learning, li2017learning, douillard2020podnet, Cha_2021_ICCV}.
In malware detection, \sys{} markedly mitigates forgetting while maintaining adaptability—achieving an average retention score improvement of $33.3\%$ and even surpassing the oracle upper bound~(full retraining~\cite{wu2023multi}) on new data.
In binary summarization, \sys{} consistently outperforms all baselines across abstraction levels and achieves more than twice the BLEU score of the prior transfer-learning approach on the critical stripped-binary task.
We further show that alternative strategies that preserve individual models, such as ensemble learning and model merging~\cite{yang2024model}, fail on the critical task of adapting to new malware and are significantly outperformed by CL methods in the summarization task.

\nobf{Contributions} 
This paper makes three main contributions:
\begin{itemize}
    \item We formalize CL under data-sensitive security constraints to address temporal and representational shifts. We identify two coupled challenges, i.e., retention loss and generalization failure, and trace both to uncontrolled model updates that fail to effectively integrate prior knowledge.
    \item We present \sys{} to address both challenges through \rcolor{controlled} forgetting without access to historical data. It enables data-free knowledge accumulation by consolidating previously trained and newly adapted models through parameter-level merging, where \rcolor{controlled} forgetting is enforced through adaptive low-rank and sparse updates.
    \item We evaluate \sys{} on two representative applications: malware detection under temporal drift and binary analysis under representation shift. \sys{} consistently outperforms existing CL methods on previous domains while maintaining or surpassing state-of-the-art on newly encountered, security-critical ones.
\end{itemize}

\section{Background and Motivation}

\subsection{Continual Learning (CL)}

\begin{figure}[t]
\vspace{-5pt}
    \centering
    \subfloat[Malware Detection]{%
    \includegraphics[width=0.49\linewidth]{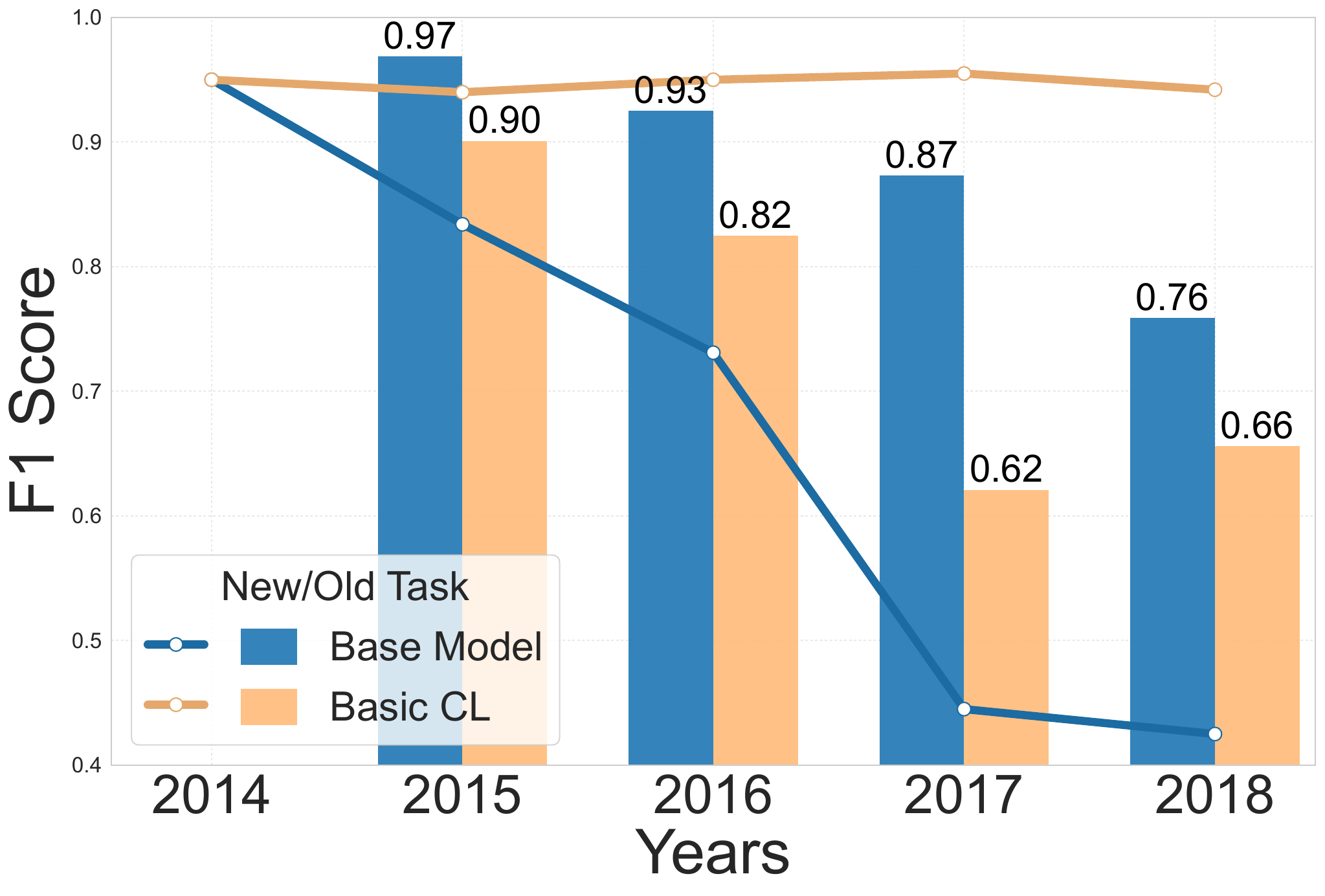}%
    \label{fig:motivate_malware}}%
    \hfill
    \subfloat[Binary Analysis]{%
    \includegraphics[width=0.49\linewidth]{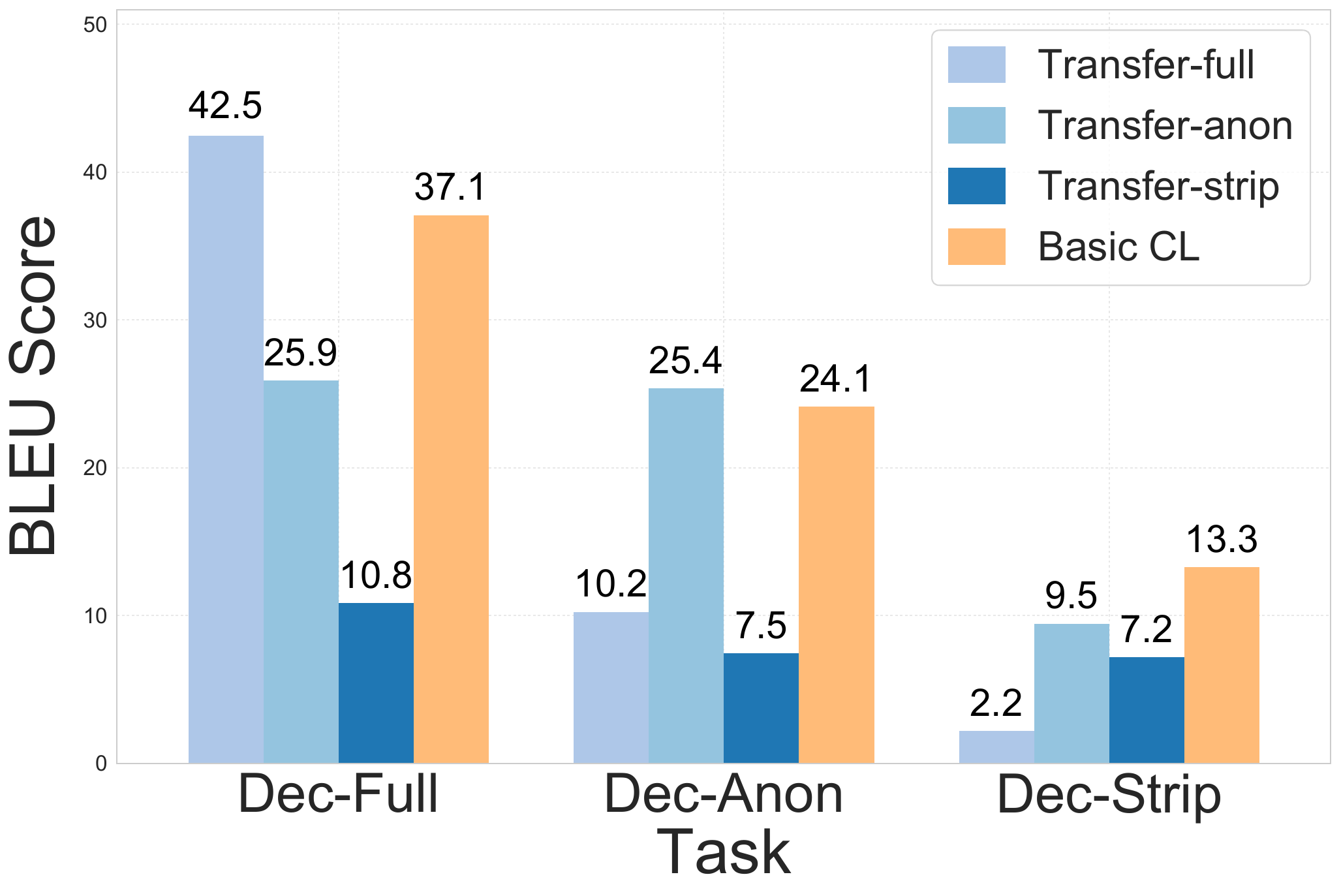}%
    \label{fig:motivate_binary}}%
    \caption{
        Examples of the benefits and insufficiencies of CL in binary security applications.
        (a)~Malware detection: continual fine-tuning mitigates temporal degradation compared to a static model trained on early data, but suffers from severe catastrophic forgetting;   
        (b)~Binary analysis: continual fine-tuning across abstraction levels outperforms transfer learning baselines in the most security-critical task, but cross-representation robustness remains challenging.
    }
    \label{fig:cl_benefits}
\end{figure}

\begin{figure*}[th]
    \centering
    \includegraphics[width=.98\linewidth]{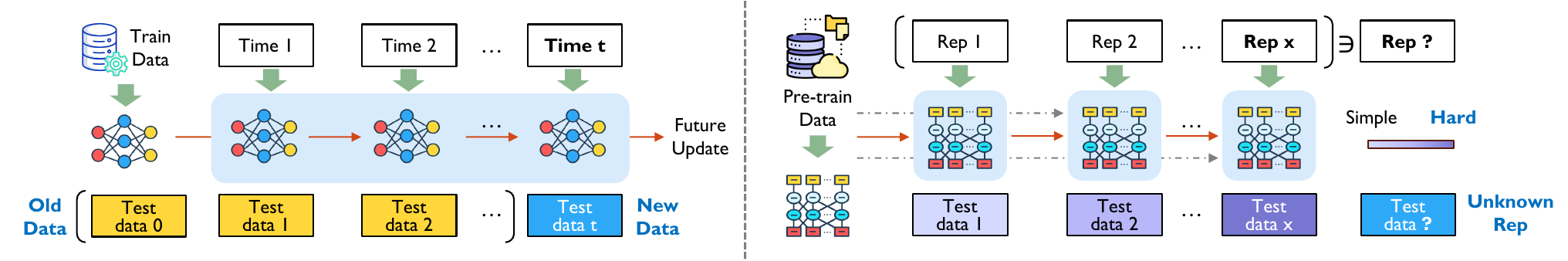}
    \caption{Continual learning for addressing temporal shift~(left) and representation shift~(right) in security applications.}
    \label{fig:security_cl_app}
\end{figure*}

In many real-world scenarios, data distributions evolve over time, requiring models to update continuously as new data arrive.  
Continual learning (CL)~\cite{de2021continual} seeks to adapt effectively to new distributions while maintaining performance on previously learned ones. 
The problems are typically categorized into three main scenarios based on how tasks are defined and evaluated~\cite{wang2024comprehensive}:
task-incremental (task identity known at inference),
class-incremental (no explicit task ID, expanding label space),
and domain-incremental (shared label space, changing input distribution).
Our work addresses the \emph{domain-incremental} setting, which is highly relevant to real-world security applications, as it directly models the challenges of temporal drift and representation shift (defined in~\autoref{sec:cl_sec_settings}). 

A common naive baseline for this setting is \textit{continual fine-tuning (CFT)}, where the model is sequentially full fine-tuned on each new dataset.  
However, this approach typically suffers from \textit{catastrophic forgetting}~\cite{guo2025persistent}, where updates on new data overwrite representations learned from previous distributions.  
Many existing CL methods are thus designed to mitigate forgetting and are typically grouped into three main categories: 
    
    \nobf{Replay-based} These methods explicitly preserve prior knowledge by storing data from previous tasks and interleaving it with new samples during training.
    Representative approaches include Experience Replay (ER)~\cite{rolnick2019experience} and iCaRL~\cite{rebuffi2017icarl}, both of which adopt selective sample replay.
    While ER relies on simple random sampling of past data, iCaRL designs a more structured strategy by retaining samples whose feature representations are closest to the class mean in the embedding space.
    
    \nobf{Regularization-based} This category constrains model updates by preserving dynamics of earlier knowledge.  
    \emph{Weight-constraint} methods often require preserving detailed information from the old model's training process. 
    They adopt common heuristics including the Fisher information (EWC~\cite{kirkpatrick2017overcoming}), path-integral contributions (SI~\cite{zenke2017continual}), and output sensitivity (MAS~\cite{aljundi2018memory}).
    In contrast, knowledge \emph{distillation-based} methods preserve the functional knowledge of the old model. For example, the classical approach LwF~\cite{li2017learning} treats the old model's logits as a distillation target, while subsequent works improve by distilling feature-level representations (PODNet~\cite{douillard2020podnet}) or adding contrastive learning (Co\textsuperscript{2}L~\cite{Cha_2021_ICCV}).
    
    \nobf{Architecture-based} These methods mitigate forgetting by modifying the network structure, such as by isolating parameters for different tasks~(PackNet~\cite{mallya2018packnet}, HAT~\cite{serra2018overcoming}) or dynamically expanding the network~(DEN~\cite{yoon2018lifelong}). Since most approaches rely on task labels at inference time to know which mask or sub-network to use, they are restricted to the task-incremental setting. Some recent works, like ResAdapt~\cite{rebuffi2017learning}, have adapted this paradigm to domain-incremental and introduces lightweight adapters.

While these paradigms show effectiveness in controlled benchmarks from computer vision and natural language processing~\cite{woo2023convnext, tan2021efficientnetv2}, they often rely on idealized assumptions such as stable task boundaries or accessible replay buffers. 

\subsection{CL for Binary Security Applications} \label{sec:motivation}

\subsubsection{Why CL is needed} \label{sec:cl_sec_settings}
Deep learning-based binary security systems operate in dynamic environments, where the underlying data and representations continuously evolve~\cite{pendlebury2019tesseract, ding2019asm2vec}.
\autoref{fig:security_cl_app} illustrates two fundamental types of this non-stationarity and the corresponding CL settings.
\textbf{a)~Temporal shift:} the underlying data distribution changes as new behaviors, threats, or software versions emerge.
In this setting, the model is sequentially trained on datasets from time~1 to~$t$, and evaluated on both previous test sets (test$_0$,\,\dots,\,test$_{t-1}$) and the current one (test$_t$).
This scenario highlights the need for \emph{temporal robustness}, i.e., maintaining detection capability as data evolve.
\textbf{b)~Representation shift:} the similar semantics varies across abstraction levels, e.g., different stages of code transformation or recovery.
Here, a pretrained model is continually trained across representations (rep$_1$,\,\dots,\,rep$_x$), where each step introduces a more complex data form.
Evaluation is performed on an incoming test set with an unknown representation, often dominated by the most degraded but security-critical form (e.g.,~rep$_x$).

To exemplify these scenarios, this paper examines two representative applications, as shown in~\autoref{fig:cl_benefits}.
The first, \emph{malware detection}, represents the well-established motivation for continual learning~\cite{jordaney2017transcend, barbero2022transcending}:  
a model trained on early-year samples experiences sharp accuracy degradation on later data, whereas CFT maintains stable detection across years.  
The second, \emph{binary summarization}, reveals a less-explored benefit that our findings highlight in recent LLM-based code understanding~\cite{al2023extending}. 
CFT across abstraction levels of decompiled binaries leads to improvements over individually tuned models on the~rep$_x$, demonstrating effective knowledge accumulation. 
Together, these examples show how CL can provide a unified mechanism for sustaining performance across temporal and representational evolution in security-critical environments. 

\subsubsection{Security-Specific Constraints on CL} \label{sec:security_constraints}
Binary security applications introduce unique constraints that make conventional CL paradigms difficult to apply directly.  
The constraints arise from two complementary perspectives: 
{\nobf{Retrospective-free data constraints}}
Due to the sensitivity of malware samples and proprietary binaries, historical data is often subject to strict sharing, access, and reuse restrictions~\cite{ren2023demistify}.
In practice, they may be maintained by different parties or under evolving access policies across training cycles, and earlier representations are frequently unavailable when one party encounters obfuscated or stripped inputs.
Moreover, large-scale security systems (e.g., cloud-based malware detection) continuously ingest massive data streams~\cite{gao2025astbar}, rendering long-term dataset storage and repeated retraining impractical.
These factors necessitate retrospective-free continual adaptation.

{\nobf{Cross-domain performance requirements}}
Binary security models must remain effective as task boundaries are ambiguous at test time~\cite{ding2019asm2vec}.  
In malware detection, the notion of `old' and `new' data naturally coexist: novel malware families continually emerge with unseen behaviors, while legacy variants persist in the wild. 
In binary analysis, decompiled programs may originate from different abstraction levels (e.g., anonymized or stripped) without metadata indicating their form. 
Such conditions demand cross-domain robustness: the capacity to preserve prior knowledge while generalizing to new, security-critical data distributions.


\subsubsection{Research Gaps and Challenges} \label{sec:security_challenges}
The constraints above shape the applicability and effectiveness of CL in security.  
Due to limited data accessibility, many conventional CL approaches are impractical.  
Replay-based strategies directly conflict with data confidentiality and retention policies, as they require storing or sampling from historical datasets.
Similarly, regularization-based methods, particularly those relying on weight constraints, often assume the availability of shared validation data or explicitly defined task boundaries to balance stability and plasticity.
For example, EWC~\cite{kirkpatrick2017overcoming} and MAS~\cite{aljundi2018memory} still depend on previous task data to compute importance metrics, while SI~\cite{zenke2017continual} requires access to historical training trajectories—an even less feasible assumption in real-world security workflows.

Even within the feasible space, achieving effective cross-domain adaptation remains challenging.  
Without access to previous or nonexchangeable data, models easily overwrite learned representations, leading to \emph{catastrophic forgetting}.  
This problem is especially evident in time-evolving malware detection, where CFT on newer yearly data, as used in prior work~\cite{continuous}, rapidly degrades the detection of earlier malware.
Our evaluation shows that, while such models maintain strong performance on recent threats, their F1-score on older samples drops below~0.7 after three years of adaptation.
At the same time, transferring knowledge to harder domains is difficult. In binary analysis, continual adaptation from decompiled normal binaries to stripped ones still struggles to recover high-level semantics lost through symbol removal.  
These observations reveal two key gaps: the need to preserve prior knowledge without data replay and to accumulate domain-invariant representations that generalize across evolving inputs.

\begin{figure}[t]
    \centering
    \includegraphics[width=.98\linewidth]{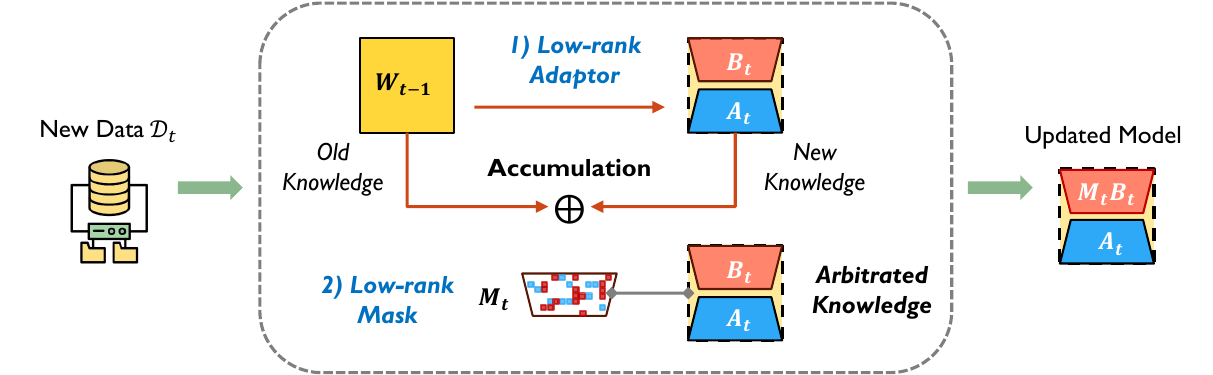}
    \caption{Design insights for accumulating old and new knowledge at each continual learning stage. Knowledge accumulation is achieved through parameter merging between the previous and newly adapted models, while interference control is ensured by applying masked low-rank updates.}
    \label{fig:design_insights}
\end{figure}

\section{Our \sys{} Method}

\subsection{Insights and Overview}
\nobf{Design Insights} 
As illustrated in~\autoref{fig:design_insights}, our method is motivated by two complementary principles that together enable retrospective-free continual learning. 
First, in the absence of old data, the previous model serves as a proxy for past knowledge, and merging its parameters with new updates provides a natural path to knowledge accumulation. 
Second, this merging must be controlled to limit interference, ensuring that new knowledge is integrated without degrading previously acquired functionality.

\noul{Accumulating Knowledge via Merging.} 
To preserve past knowledge without data replay, we draw inspiration from model merging~\cite{wortsman2022model}, where the parameters of multiple simultaneously trained models are combined to achieve strong multi-task performance. 
We adapt this idea to continual learning by incrementally merging each new update with the existing model, treating the previous model as a proxy for past data.
This consolidation enables knowledge from evolving distributions to be integrated sequentially without access to earlier samples. 
Let $W \in \mathbb{R}^{d_{\text{in}}\times d_{\text{out}}}$ denote a layer weight of the base model. 
Formally, after $T$ rounds of adaptation, the accumulated weight is
$
W_T = W_0 + \sum_{t=1}^T \Delta W_t.
$

\noul{From Full Parameters to Localized Updates.} 
To make merging more controlled, we adapt the LoRA~\cite{hu2022lora} perspective by restricting each update to a low-rank decomposition of the weight matrix. 
The original form of a LoRA update is expressed as
$
\Delta W = A B,
A \in \mathbb{R}^{d_{\text{in}}\times r},
B \in \mathbb{R}^{r \times d_{\text{out}}}, r \ll d_{\text{in}},
$
where $A$ is the down-projection and $B$ is the up-projection matrix. 
We adopt this setup for continual learning: new knowledge is localized into a compact subspace rather than distributed across the full parameter space, thereby reducing the chance of interference. 
To further refine this localization, the update is constrained on $B$ and a real-valued mask $M$ is introduced to selectively filter its components. 
Thus, for round $t$, the resulting update is $\Delta W_t = A_t (M_t \odot B_t)$.



\SetAlCapNameFnt{\small}
\SetAlCapFnt{\small}
\begin{algorithm}[t]
\DontPrintSemicolon
\caption{Continual Learning with Low-Rank \\ Updates and Adaptive Keep \& Gain Arbitration}
\small
\label{alg:continual_unified_kg}
\DontPrintSemicolon
\KwIn{Base model $\theta_{\text{base}}$ with weights in $\mathbb{R}^{d_{\text{in}} \times d_{\text{out}}}$, 
      dataset stream $\{\mathcal{D}_t^{\text{train}},\mathcal{D}_t^{\text{valid}}\}_{t=1}^N$, 
      where $\mathcal{D}_t = \{(x_i,y_i)\}$ with $x_i \in \mathbb{R}^{d_{\text{in}}}$ and $y_i \in \mathcal{Y}$}
\KwOut{Merged model $f_{\text{merge}}$ after $N$ tasks}

\textbf{Init:} $\theta_{\text{prev}} \leftarrow \theta_{\text{base}}$.\;
\For{$t=1$ \KwTo $N$}{
  \tcc{(1) Low-rank adaptation on current task}
  Draw $A_t \in \mathbb{R}^{d_{\text{in}}\times r}$ with i.i.d.\ entries of zero mean and finite variance $\sigma^2 > 0$.\;
  Init $B_t \in \mathbb{R}^{r \times d_{\text{out}}} \leftarrow 0$.\; 
  \textbf{Update} $B_t$ by minimizing \textsc{ModelLoss}$(\mathcal{D}_t^{\text{train}})$ validated on $\mathcal{D}_t^{\text{valid}}$. \Comment*[r]{freeze $\theta_{\text{prev}}$ and $A_t$}
  Define teacher models: $f_{\text{old}} \gets \theta_{\text{prev}}$, 
  $f_{\text{new}} \gets \theta_{\text{prev}} + \text{scaled }(A_t B_t)$.\;

  \tcc{(2) Learn mask for knowledge merging}
  Init $M_t \in \mathbb{R}^{r \times d_{\text{out}}} \propto B_t$. \Comment*[r]{mask on $B_t$}
  \Repeat{convergence}{
    Sample $(x,y)\!\sim\!\mathcal{D}_t^{\text{train}}\!\cup\!\mathcal{D}_t^{\text{valid}}$.\;
    \tcp{Model predictions}
    $p_o=f_{\text{old}}(x)$, $p_t=f_{\text{new}}(x)$;\; 
    $p_s=(\theta_{\text{prev}} + \text{scaled }(A_t(M_t\odot B_t)))(x)$.
    

    \tcp{Calibrated true-class confidence}
    $\mathcal{C}_o=(c_o, \tau_o) \leftarrow \textsc{Conf}(p_o,y;f_{\text{old}})$;\;
    $\mathcal{C}_t=(c_t, \tau_t) \leftarrow \textsc{Conf}(p_t,y;f_{\text{new}})$.\;

    \tcp{Keep and Gain arbitration}
    $T^\star \leftarrow \textsc{Teach}(p_o,p_t;\,\mathcal{C}_o,\mathcal{C}_t)$, \;
    $w^\star \leftarrow \textsc{SampleWeight}(\mathcal{C}_o,\mathcal{C}_t)$; \;
    $\mathcal{L}_{\text{merge}} \leftarrow \mathbb{E}\big[w^\star \cdot \textsc{Arbitrate}(p_s \,\|\, T^\star)\big]$.\;

    \tcp{Supervision \& mask sparsity}
    $\mathcal{L}_{\text{sup}}=\textsc{ModelLoss}(p_s,y)$;\;
    $\mathcal{R}(M_t)=\textsc{SparseGroupLasso}(M_t)$. \; 
    \textbf{Update} $M_t$ by minimizing total loss $\mathcal{L} \leftarrow \mathcal{L}_{\text{merge}} + \eta \mathcal{L}_{\text{sup}} + \mu\,\mathcal{R}(M_t)$.\;
  }

  \tcc{(3) Merge into the running model}
  $\theta_{\text{prev}} \leftarrow
  \theta_{\text{prev}} + \text{scaled }(A_t(M_t\!\odot\!B_t))$. \Comment*[r]{current merged}
}
\textbf{Return} $f_{\text{merge}} \leftarrow \theta_{\text{prev}}$.\;
\end{algorithm}

\nobf{Method Overview} 
\autoref{alg:continual_unified_kg} summarizes the overall procedure. 
Building on the insights of localized low-rank updates and cumulative merging, each task-specific adaptation is restricted to the projection matrix $B_t$~(Lines~3–7), and the resulting masked update is incorporated into the running model~(Line~21). 
Arbitration~(Lines~8–20) then complements this foundation by regulating how new knowledge is integrated during merging: the learned masks limit parameter overlap across tasks, while confidence-guided transfer ensures that reliable past predictions are preserved.

\noul{\rcolor{Controlled} Forgetting.} 
The combination of structural control and arbitration \rcolor{introduces a soft bound on update-induced forgetting.} 
Formally, let the effective update energy at round $t$ be $E_t := \|M_t \odot B_t\|_F$, which measures the Frobenius norm of the masked low-rank expansion actually merged. 
If $A$ is an $\varepsilon$-approximate isometry, then the average prediction drift introduced by the $t$-th merge satisfies
\begin{equation}
\begin{aligned}
& \mathbb{E}_{\mathcal{D}_{\le t-1}}
\!\big[\|p_t-p_{t-1}\|_2\big]
\;\le\; \\
& O\!\Bigg(
\sqrt{\tfrac{r}{d_{\mathrm{in}}}} \sum_{s=1}^{t} E_s
+ \rho \sum_{s=1}^{t} E_s
+ \sqrt{\delta}
+ \varepsilon
\Bigg),
\end{aligned}
\label{equ:bound}
\end{equation}
where $\mathcal{D}_{\le t-1}$ denotes the union of past distributions and $p$ is the predictive probability distribution of the model. 


In the following, we explain different components of this forgetting bound. 
The first term, which scales with \(\sqrt{r/d_{\mathrm{in}}}\), reflects structural interference control through low-rank adaptation and is analyzed in \autoref{sec:method_lowrank}.
The latter two terms, involving mask overlap $\rho$ and arbitration tightness $\delta$, arise from the arbitration stage and are examined in \autoref{sec:method_arbitration}.


\subsection{Low-rank Model Adaptation} \label{sec:method_lowrank}

This section analyzes how the structural design of our update mechanism mitigates task conflict during continual learning. 
To make this concrete, we first describe the update–merge formulation, showing how only $B_t$ is trained and integrated into the model while the random projection $A$ remains fixed. 
We then establish two theoretical properties~(proofs in~\aref{appendix:proofs}) to explain the reduced knowledge interference.

\noul{Update–Merge on $B_t$.}
In our design, all task-specific knowledge is concentrated in the low-rank coefficients $B_t$, while both the base model $\theta_{\text{prev}}$ and the projection matrix $A$ remain fixed. 
For each round $t$, the expansion matrix $B_t$ is optimized to minimize the model loss on new data~(Line~5). 
The mask $M_t$ is then initialized from the drift captured in $B_t$~(Line~7) and refined subsequently during arbitration to regulate how the update is merged. 
Let $\theta$ denote the parameters of the base model, and consider a representative weight matrix $W \in \mathbb{R}^{d_{\text{in}} \times d_{\text{out}}}$ within $\theta$, such as a projection in an MLP or Transformer block. The merged weight after $T$ updates is
\begin{equation}
W_T = W_0 + \sum_{t=1}^T A_t (M_t \odot B_t).
\end{equation}
Specifically, the low-rank update for task $t$ follows our mask-based LoRA form, where only $B_t$ (for adaptation) and $M_t$ (for arbitration) are trainable.
This setup ensures that interference across tasks arises solely from overlaps among the matrices $\{B_t\}$, and thus inter-task variability is entirely confined to the $r \times d_{\text{out}}$ coefficient space.



\noul{Effect of Low Rank.}
By projecting the update onto an $r$-dimensional manifold, the `surface area' for potential interference with previous knowledge is reduced from $d_{\text{in}}$ to $r$.




\begin{proposition}[Low-rank reduces expected interference]
\label{prop:lowrank}
Let $g_{\text{old}} = \nabla_\theta \mathcal{L}_{\text{old}}(\theta_{\text{prev}})$ denote the gradient of a past task at the current model. 
For any old-task gradient $g_{\text{old}}$ and update $\Delta W_t = A_tB_t$,
\begin{equation}
\mathbb{E}\!\left[\frac{\langle g_{\text{old}}, \Delta W_t\rangle}{\|g_{\text{old}}\|_F \|\Delta W_t\|_F}\right]
\;\leq\; O\!\left(\sqrt{\tfrac{r}{d_{\text{in}}}}\right).
\end{equation}

Thus, the expected cosine similarity between new updates and old-task gradients
scales with \(\sqrt{r/d_{\mathrm{in}}}\), making destructive interference less
likely as long as \(r \ll d_{\mathrm{in}}\).
\end{proposition}

\noul{Effect of Freezing $A$.}
The matrix $A$ is randomly initialized with zero mean and positive variance~(Line 3) and remains frozen throughout training~(Line 5). 
By fixing $A$, we leverage a high-dimensional phenomenon where random directions are naturally perpendicular to one another, resulting in approximate orthogonality in high-dimensional regimes. 



\begin{proposition}[Approximate orthogonality with frozen random $A$]
\label{prop:frozenA-orth}
Let $A_s, A_t \in \mathbb{R}^{d_{\text{in}}\times r}$ be independent random matrices
for distinct tasks $s\neq t$, whose entries are i.i.d.\ with zero mean and positive variance. 
Assume the entry distribution is sub-Gaussian.
Let $\widetilde{B}_s := B_s \odot M_s$ and $\widetilde{B}_t := B_t \odot M_t$, and define
$\Delta_s := A_s\widetilde{B}_s$ and $\Delta_t := A_t\widetilde{B}_t$.
If $r = o(\sqrt{d_{\text{in}}})$, then as $d_{\text{in}}\to\infty$, 
\begin{equation}
\frac{\langle \Delta_s,\Delta_t\rangle_F}
{\|\Delta_s\|_F\,\|\Delta_t\|_F}
\xrightarrow[]{\mathrm{p}} 0 .
\end{equation}

Hence, given task-specific updates $\Delta_s$ and $\Delta_t$, freezing the random projections $A$ ensures that task directions are asymptotically orthogonal. 
As the cross-task interaction (cosine similarity) becomes negligible in high-dimensional regimes, multiple updates can be merged into a single model without significant interference.
\vspace{-5pt}
\end{proposition}

\noul{Remarks.}
By restricting updates and merges exclusively to $B_t$, the model localizes all task-specific knowledge to a controlled subspace. 
The interference is bounded by \(\sqrt{r/d_{\mathrm{in}}}\), while the frozen random projection $A$ induces approximately orthogonal updates across tasks.

\rcolor{Note that our analysis is conducted at the \emph{parameter level}. Although the full model may be nonlinear, adaptation is still realized through structured perturbations to trainable modules, e.g., linear projections and feed-forward matrices in Transformers. 
Under standard local linearization assumptions, their effect is governed by their geometry in parameter space, making our analysis meaningful beyond purely linear models~(see \aref{appendix:model_level}). The result should therefore be understood as a \emph{local controlled-forgetting guarantee}, not a strict global guarantee in highly non-convex function space.}


\subsection{Confidence-driven Knowledge Arbitration} \label{sec:method_arbitration} 
We next analyze how the arbitration stage complements low-rank adaptation by regulating both functional preservation and parameter overlap. 
While structural design restricts interference to a subspace, naive merging can still introduce conflicts when updates overlap or when old-task predictions are altered. 
To address this, arbitration introduces two key mechanisms that correspond to the residual terms in \autoref{equ:bound}: 
1)~a confidence-driven knowledge distillation that preserves predictions in high-confidence regions of teachers to control functional tightness~$\delta$, and 
2)~sparse mask learning with group regularization, which reduces overlap between updates and controls the mask overlap ratio~$\rho$.

\begin{figure}[t]
    \centering
    \includegraphics[width=.98\linewidth]{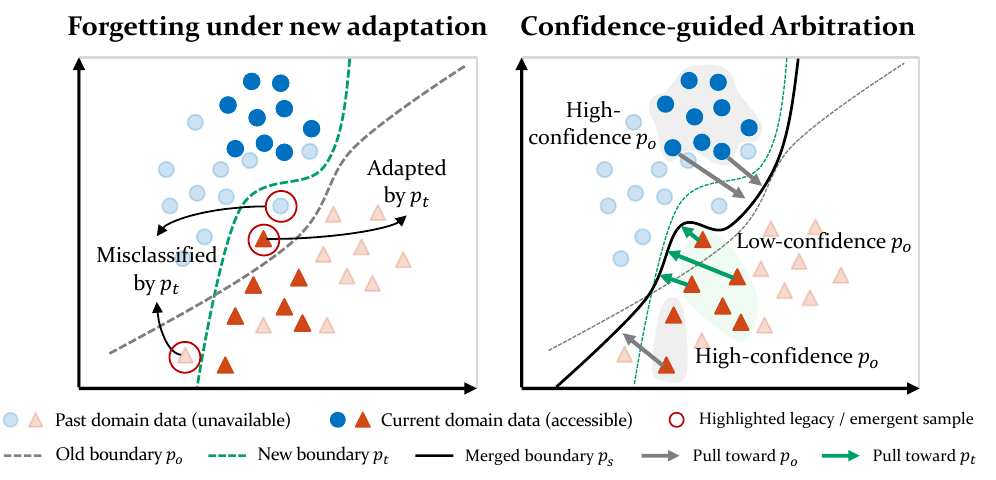}
    \captionsetup{labelfont={color=tcolor}, textfont={color=tcolor}}
    \vspace{-8pt}
    \caption{Intuitive examples of how confidence-guided arbitration pulls the decision boundary to preserve legacy knowledge while adapting to emergent knowledge.}
    \label{fig:arbitration_intuition}
\end{figure}

\noul{Confidence-guided Arbitration.}
For a training sample $(x, y)$ from the current task, let $p(\cdot)=f(x)$ denote some model output distribution. 
The student prediction $p_s$ is trained to reconcile the outputs of two teachers~(Lines~10--11): 
the previous model $p_o=f_{\text{old}}(x)$ and the adapted model $p_t=f_{\text{new}}(x)$. 

\rcolor{As shown in~\autoref{fig:arbitration_intuition}, new adaptation may derive a decision boundary that overfits to current-domain data, causing legacy knowledge encoded by the old model to be forgotten, as highlighted legacy samples become misclassified in the left panel.}
Since only data from the current domain are accessible, the old model serves as the sole proxy for past knowledge. 
Arbitration therefore \emph{prioritizes the old model} \rcolor{when its prediction is reliable to maintain stability, as shown in the right panel.}
When the old model is confident on the true class, the sample likely lies within regions where \rcolor{$p_o$ has a large decision margin and} encodes reliable legacy knowledge. 
Enforcing consistency with $p_o$ in these cases prevents overwriting reliable representations\rcolor{, thereby correctly classifying the highlighted legacy samples.} 
Only when the old model is less certain does arbitration shift toward the new model $p_t$ to strengthen knowledge acquisition\rcolor{, which keeps the highlighted emergent sample correctly classified.}

To this end, we implement arbitration with four components: calibrated confidence estimation, teacher selection, sample reweighting, and distillation to the student.
The true-class confidence $c$ is computed as the model’s softmax probability for the ground-truth class $y$. To improve cross-class comparability, an optional conformal percentile threshold $\tau\in[0,1]$ can also be computed based on validation data confidences~(Lines~12–13). 
We follow standard practices from prior work~\cite{barbero2022transcending, virk2025calibration} for this optional calibration, and a prediction is then regarded as confident if $c > \tau$.
The resulting confidence pairs $\mathcal{C}_o=(c_o,\tau_o)$ and $\mathcal{C}_t=(c_t,\tau_t)$ are then used to determine which teacher to follow for each sample.

Formally, the arbitration loss guided by confidence to enforce consistency with different models~(Lines~14–15) is
\begin{equation}
\mathcal{L}_{\text{merge}} 
= \mathbb{E}[w^\star \cdot \textsc{Arbitrate}(p_s \,\|\, T^\star)],   
\label{equ:merge_loss}
\end{equation}
where $T^\star$ denotes the arbitration target derived from teacher predictions $\{p_o, p_t\}$, 
$w^\star$ is a sample-wise weight proportional to teacher reliability, 
and $\textsc{Arbitrate}(\cdot)$ measures the divergence (e.g., KL or MSE) between the student prediction and the selected or blended teacher outputs. 

\nobf{Arbitration Modes} 
We instantiate $T^\star$ and $w^\star$ under two confidence-based modes.

\noindent
    1)~\emph{Hard arbitration.} 
    A discrete confidence rule selectively activates each teacher based on its own reliability:
    \begin{equation}
    T^\star = p_o \bm{1}[c_o\!\ge\!\tau_o]
            + p_t \bm{1}[c_o\!<\!\tau_o,\, c_t\!\ge\!\tau_t],
    \quad w^\star=1.
    \end{equation}
    Only confident and correct teachers contribute to supervision, while low-confidence predictions are excluded from the merge loss. 
    This selective strategy enforces reliability at each step: the old model preserves established knowledge when trustworthy, and the new model contributes only when it confidently improves on unseen or shifted concepts.  
    This mode is typically favored when precise control over forgetting is required or when label noise is minimal.

\noindent
    2)~\emph{Soft arbitration.} 
    A continuous gate produces a convex combination of teachers using a smooth old-priority gate:
    \begin{equation}
    g = \sigma\!\left({c_o - \tau_o}\right) \in [0,1],
    \qquad
    T^\star = g\,p_o + (1-g)\,p_t.
    \end{equation}
    The gate $g$ increases monotonically with old-model confidence, yielding a smooth transition from the old to the new teacher as confidence decreases. 
    An optional weighting term $w^\star$ can further modulate each sample’s contribution based on confidence or disagreement, for example:
    $w^\star \propto \max(c_o, c_t)^{\beta}\big(1 + \lambda\,\|p_o - p_t\|_1\big),$
    where $\beta$ and $\lambda$ control the emphasis on teacher reliability and disagreement, respectively.
    This formulation allows both teachers to contribute proportionally, avoiding abrupt supervision shifts, and is advantageous when representations evolve progressively, such as in feature-level or generative tasks.


    

The arbitration loss is integrated into the overall objective (Line~19), along with supervision and sparsity regularization, driving a sparse mask learning process for knowledge accumulation, which we introduce next.

\noul{Sparse Mask Learning.}
A mask $M_t$ is learned over the low-rank adapter $B_t$ to control parameter overlap across tasks. 
We impose a Sparse-Group-Lasso regularization that promotes both element-wise and structured sparsity:
\begin{equation}
\mathcal{R}(M_t)
= \|M_t\|_1 + \lambda \sum_b \|M_{t,b}\|_2,
\label{equ:sparsity}
\end{equation}
where $\lambda$ balances fine-grained and group-level sparsity. 
The first term enforces element-wise pruning for compactness, while the second term encourages group-level disjointness, with each group $b$ corresponding to a rank column in the LoRA adapter $B_t$. 
This structured sparsity enhances interpretability by identifying which low-rank components are activated and isolating task-specific subspaces.
Finally, the mask is optimized jointly with arbitration through:
\begin{equation}
\mathcal{L}
= \mathcal{L}_{\text{merge}}
+ \eta\,\mathcal{L}_{\text{sup}}
+ \mu\,\mathcal{R}(M_t),
\label{equ:overall_loss}
\end{equation}
where $\mathcal{L}_{\text{sup}}$ denotes the supervised loss (e.g., cross-entropy) on the current task, which anchors the optimization to ground-truth labels to avoid drifting toward purely self-consistent representations. 
The optimization proceeds until convergence, after which the resulting masked adapter $A_t(M_t \odot B_t)$ is merged into the running model $\theta_{\text{prev}}$ (Line~21). 


\noul{Remarks.}
Confidence-driven knowledge arbitration clamps functional drift where the old model is reliable and constrains structural interference via the sparse mask, providing safeguards for stable and interpretable \emph{knowledge retention}. 
Meanwhile, it guarantees plasticity on new data by releasing these constraints when the old model is uncertain, enabling effective adaptation to novel concepts.
Formally, the expected predictive drift on the current distribution $\mathcal{D}_t$ satisfies 
\begin{equation}
   \mathbb{E}_{\mathcal{D}_t}\!\left[\|p_t - p_{t-1}\|_2\right] \ge C(1-\rho)E_t, 
\end{equation}
where  $C>0$ is a model-specific constant reflecting sensitivity to drift, complementing the earlier \rcolor{soft} forgetting bound to illustrate control over \emph{knowledge adaptation}.


\section{\sys{} Deployment} \label{sec:deploy}
We demonstrate the deployment of \sys{} under two representative CL scenarios in binary security. 

\nobf{Temporal Drift~(Malware Detection Example)}
This scenario reflects the continual evolution of malicious behaviors over time, where each temporal segment introduces new families or variants unseen in earlier datasets. 
Older samples remain critical for preserving established family patterns and recurring code reuse, while newer samples are the most security-critical, often embedding novel evasion strategies. 
Therefore, \sys{} ensures backward-compatible detection while maintaining adaptability to emerging threats.
Because malware datasets are typically well-annotated, the \emph{hard arbitration} mode is adopted, enforcing discrete teacher selection for precise control over forgetting.
Formally, the arbitration in~\autoref{equ:merge_loss} operates on probability distributions via a KL divergence:
$
\sum_{i} p_s(i) \log \frac{p_s(i)}{T^\star(i)}
$,
where $p_s$ denotes the student’s output logits and $T^\star$ is that from the selected teacher. 

\nobf{Representation Shift~(Binary Analysis Example)}
In binary analysis, executables appear at different abstraction levels, from normally compiled to stripped versions. 
While high-level representations provide richer semantics, stripped binaries are the most security-critical, often being the only artifacts available during incident response. 
\sys{} sustains robustness across such heterogeneous representations by transferring structural knowledge between them. 
Because the model is generative and the summary ground truth is often noisy~(e.g., extracted from comments), the \emph{soft arbitration} mode is adopted to enable confidence-weighted blending. 
It is performed in latent space of the model via a mean squared error:
$
\|p_s - T^\star\|_2^2
$,
where $p_s$ denotes pooled latent representations of the student model, and $T^\star$ is the confidence-weighted latent interpolation between that of teacher models.

We use these deployment setups for evaluation and provide more implementation details of our method in~\aref{appendix:our_implementation}.

\section{Evaluation} \label{sec:evaluation}

\subsection{Experimental Setup} \label{sec:eva_setup}

\nobf{Dataset and Model}
For the malware detection task, we adopt the dataset from Transcendent~\cite{barbero2022transcending}, which spans five years~(from 2014 to 2018) and mitigates common sources of temporal and spatial bias.
It contains 232,848 benign and 26,387 malicious Android applications collected from AndroZoo.\footnote{AndroZoo: \url{https://androzoo.uni.lu}}
Following prior work, we extract the widely used Drebin features~\cite{arp2014drebin} and employ a multilayer perceptron~(MLP) classifier as the backbone model~\cite{grosse2017adversarial}.
For binary analysis, we leverage the CAPYBARA~\cite{al2023extending} dataset, which provides function–documentation pairs extracted from decompiled binaries.
The decompiled code is organized into three subsets to represent progressively reduced levels of semantic information that emulate real-world obfuscation~(\emph{decompiled-}): \emph{full}, \emph{anonymized}, and \emph{stripped}.\footnote{%
Subset names in its original paper: \emph{Decompiled}, \emph{Demi-Stripped}, and \emph{Stripped}.
We adopt clearer names to improve interpretability.
Specifically, the anonymized variant (originally Demi-Stripped) removes all identifiers from the decompiled code and replaces them with generic placeholders.}
The dataset contains 215K aligned and documented decompiled functions, while only 14K documented samples are available for the stripped subset, underscoring the difficulty of recovering meaningful semantics from heavily stripped binaries.
Following the dataset's accompanying BinT5 model design, we employ the same CodeT5~\cite{wang2021codet5} backbone to ensure consistency.
\rcolor{Within each temporal split and abstraction subset, data are partitioned with a consistent 8:2 train/test split throughout all CL experiments.}

\nobf{CL Baselines}
We compare \sys{} with five representative continual learning methods, encompassing the primary strategies compatible with our challenging replay-free and task-agnostic setting.
1)~\textbf{CFT}: the naive sequential training baseline that fine-tunes the model on each new task without any mechanism to mitigate forgetting.
2)~\textbf{LwF}~\cite{li2017learning}: a logit-distillation method that regularizes the new model’s training using the previous model’s predictions as soft targets.
3)~\textbf{PODNet}~\cite{douillard2020podnet}: a feature-distillation approach that constrains both the output and intermediate feature representations of the new model to remain consistent with those of the previous one.
4)~\textbf{Co\textsuperscript{2}L}~\cite{Cha_2021_ICCV}: a contrastive learning method that designs an instance-relationship distillation loss that preserves pairwise similarity structures from the previous model.
5)~\textbf{ResAdapt}~\cite{rebuffi2017learning}: an architecture-based method that retains the base weight and expands the network with residual adapters for new tasks.
The distillation-based baselines (2$\sim$4) may share similar concepts to our knowledge arbitration but differ fundamentally in mechanism: these methods regularize new model training, whereas \sys{} performs model consolidation through mask-based merging.

For reference, two additional models are included as anchors. 1)~\emph{Base}: a static model trained only on the first-year data for malware detection or the stripped-binary subset for binary analysis. 2)~\emph{Oracle}: a joint training model using all available data at each period for malware detection or across all representation levels for binary analysis.
Implementation details of these baselines are provided in~\aref{appendix:cl_baseline}, with comparative results reported in~\autoref{sec:eva_cmp_cl} for malware detection and~\autoref{sec:eva_cmp_cl_rep} for binary analysis.

\nobf{Aggregation Baselines}
As an alternative to continual learning, aggregation-based methods maintain multiple expert models and combine their knowledge at prediction or parameter level instead of performing sequential adaptation.
We include two ensemble approaches: \emph{hard voting}, which selects the majority prediction among experts, and \emph{soft averaging}, which aggregates their probability outputs.
Beyond ensembles, four representative model-merging techniques are considered. \emph{Weight Averaging}~\cite{wortsman2022model} directly averages parameters across models; \emph{Task Arithmetic}~\cite{ilharco2022editing} transfers task-specific knowledge via linear operations on task vectors; \emph{TIES-Merging}~\cite{yadav2023ties} resolves conflicting parameters through sign-based majority rules; and \emph{AdaMerging}~\cite{yang2023adamerging} learns adaptive layer-wise merge weights.
Implementation details are provided in~\aref{appendix:aggregate_baseline}, and their comparative results across both applications are presented in~\autoref{sec:eva_cmp_ensemble}.



\begin{figure}[t]
    \centering
    \includegraphics[width=.92\linewidth]{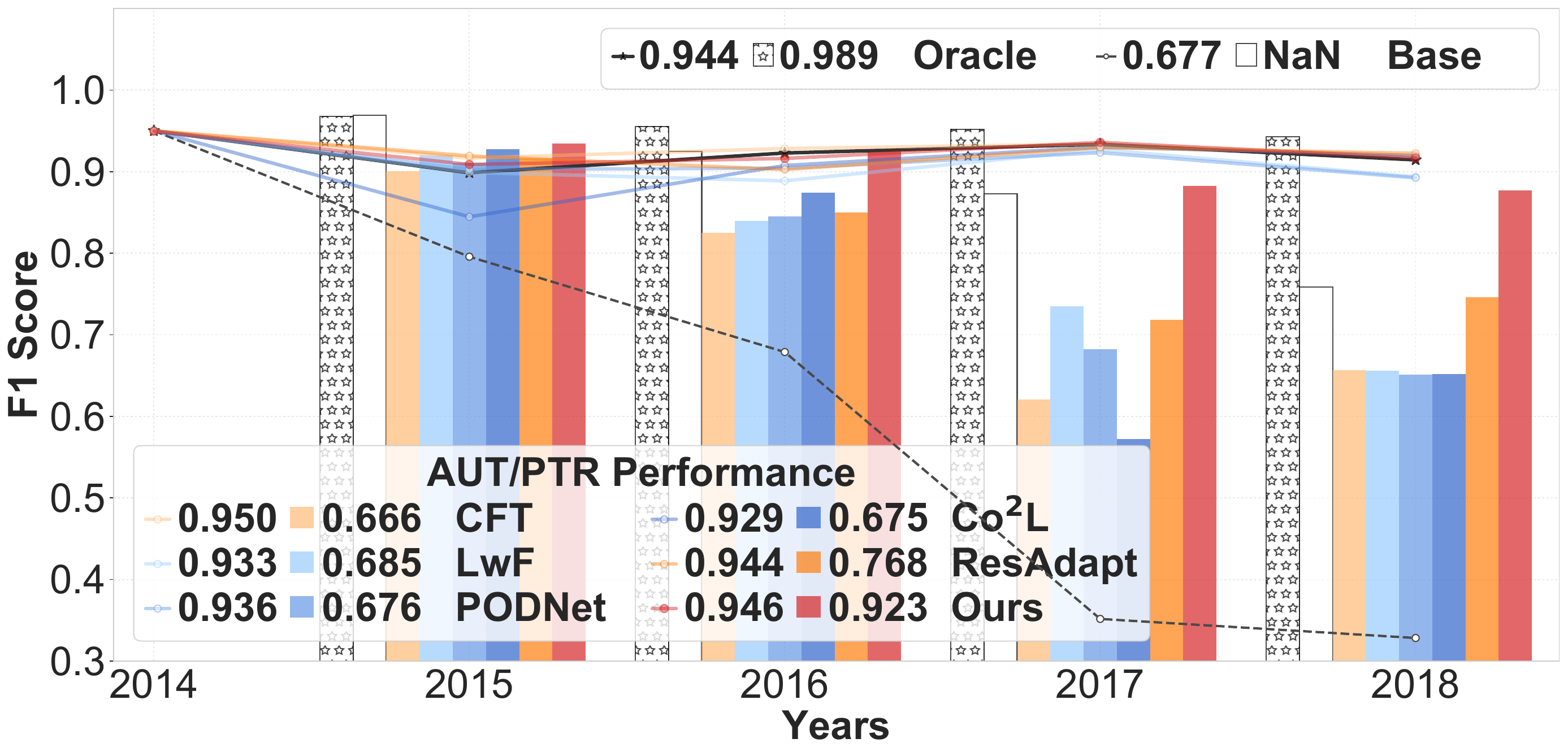}
    \vspace{-1em}
    \caption{Cumulative CL comparison in malware detection.}
    \label{fig:cmp_cl_malware}
\end{figure}

\begin{table}[t]
    \centering
    \caption{{Year-wise CL comparison in malware detection: \rcolor{per-step} F1-scores on old and new test data \rcolor{(bar and line values)}.}}
    \vspace{-8pt}
    \begin{smalltabularx}{\linewidth}{lCCCCCCCC}
    \toprule
           & \multicolumn{2}{c}{\textbf{2015}} & \multicolumn{2}{c}{\textbf{2016}} & \multicolumn{2}{c}{\textbf{2017}} & \multicolumn{2}{c}{\textbf{2018}} \\ \cmidrule(lr){2-3} \cmidrule(lr){4-5} \cmidrule(lr){6-7} \cmidrule(lr){8-9}
           & Old         & New        & Old         & New        & Old         & New        & Old         & New        \\ \midrule
    Base   &      0.969       &      0.834      &       0.925      &      0.731      &     0.873        &     0.445       &       0.759      &      0.425      \\
    Oracle &      0.967       &      0.924      &       0.955      &      0.945      &      0.952       &      0.954      &       0.943      &       0.938     \\ \midrule
    CFT    &       0.901      &      0.940      &       0.825      &      \textbf{0.950}      &     0.621        &      \textbf{0.955}      &      0.656       &      0.942      \\
    LwF~\cite{li2017learning}    &      0.920       &      0.924      &       0.840      &      0.915      &      \textbf{0.735}       &      0.949      &      0.656       &      0.919     \\
    PODNet~\cite{douillard2020podnet} &      0.911       &      0.927      &       0.845      & 0.929           &       0.682      &      0.945      &       0.651      &       0.918     \\
    Co\textsuperscript{2}L~\cite{Cha_2021_ICCV}   &       \textbf{0.927}      &      0.877      &       \textbf{0.875}      &       0.932     &      0.572       &      0.951      &      0.652       &      0.942      \\
    ResAdapt~\cite{rebuffi2017learning} &      0.916       &      \textbf{0.942}      &      0.850       &     0.927       &      0.718       &      0.951      &      \textbf{0.746}       &       \textbf{0.945}     \\ \midrule
    \textbf{\sys{}} &      \textbf{0.935}       &      \textbf{0.933}      &      \textbf{0.926}       &      \textbf{0.939}      &          \textbf{0.882}   &      \textbf{0.956}      &       \textbf{0.877}      &      \textbf{0.940}     \\ \bottomrule
    \end{smalltabularx}
    \label{tbl:cmp_cl_malware}
\end{table}

\subsection{Effectiveness on Temporal Drift} \label{sec:eva_cmp_cl}


We first evaluate \sys{} on temporal drift using the long-standing benchmark of Android malware detection, comparing its performance against representative CL baselines.

\nobf{Evaluation Metrics}
Following standard practice, we report the F1-scores on the latest test data~(new threats) and all previously encountered test sets~(old threats).
To quantify overall adaptability across time, we use the AUT metric~\cite{pendlebury2019tesseract}, defined as:
$
\text{AUT}(N, f) = \frac{1}{N-1} \sum_{k=1}^{N-1} \frac{f(k+1) + f(k)}{2}
$,
where $N$ is the number of evaluation slots, and $f(k)$ denotes the F1-score at time $k$.
A perfect classifier with no temporal degradation achieves an AUT of~1.
To assess retention of previously acquired knowledge, we define the Past-Task Retention~(PTR) score:
$
\text{PTR}_t = \frac{1}{t-1} \sum_{s=1}^{t-1} 
\frac{a_{t,s}}{a^{\max}_{\le t, s}}
$,
where $a_{t,s}$ is the \rcolor{F1-}score on the $s$-th test set after learning up to time~$t$, and $a^{\max}_{\le t, s}$ denotes the highest \rcolor{F1-}score ever achieved on that set.
PTR is consistently reported at the final time step to summarize forgetting accumulated over the entire sequence. 
Note that PTR is computed only for continual models that undergo sequential updates across years and not applicable to the static base model, which serves as a fixed reference for absolute performance.

\nobf{Result Analysis}
We present the cumulative performance of \sys{} and all CL baselines across years in~\autoref{fig:cmp_cl_malware}, with detailed per-year results on both historical (old) and current (new) test sets shown in~\autoref{tbl:cmp_cl_malware}.
All CL baselines exhibit reasonable adaptability to newly emerging threats but experience substantial forgetting of earlier malware samples.
The naive CFT baseline achieves strong new-year performance but exhibits the worst retention, with its PTR dropping to only~$0.666$ after the four-year sequential updates.
In contrast, \sys{} improves the retention score by~$38.6\%$ over CFT while preserving comparable adaptation to new data.
Distillation-based methods such as LwF, PODNet, and Co\textsuperscript{2}L yield moderate improvements in retention (exceeding CFT by~$1.3\sim2.9\%$) thanks to their knowledge-regularization losses, yet this comes at the cost of adaptability~(with AUT values decreasing by about~$1.8\%$ on average).
Among these, PODNet favors adaptation, whereas LwF favors retention, but their overall performance remains similar, suggesting that regularization alone is insufficient to balance.
The architecture-based method ResAdapt achieves the highest retention among the baselines, but at the cost of network expansion, which increases training cost and model size; \sys{} still surpasses it by~$20.2\%$ in PTR while retaining competitive AUT.

Overall, \sys{} achieves the best balance between retention and adaptability. 
It improves the PTR by  $33.3\%$ across all baselines on average while maintaining or exceeding their AUTs.
Being conceptually aligned with distillation-based methods during knowledge arbitration, \sys{} attains its clear advantage through low-rank model consolidation rather than sample-level knowledge transfer, resulting in superior stability in preserving old knowledge each year.
Notably, \sys{} even outperforms the oracle in most yearly \rcolor{new-test} evaluations and the cumulative AUT metric.
This suggests that \rcolor{controlled} forgetting not only mitigates catastrophic forgetting but also enhances \rcolor{specialization. This phenomenon also exists in other applications, for which we provide a thorougher discussion in~\autoref{sec:discussion}.}





\begin{table*}[t]
\centering
\caption{Comparison with model-preserving baselines across malware detection and binary analysis.}
\vspace{-8pt}
\setlength{\aboverulesep}{1.1pt}
\setlength{\belowrulesep}{1.1pt}
\begin{threeparttable}
\begin{smalltabularx}{.9\linewidth}{p{11em}|CCCCCC|CCCC} 
\toprule
& \multicolumn{6}{c|}{\textbf{Malware Detection}} & \multicolumn{4}{c}{\textbf{Binary Analysis}} \\
\cmidrule(lr){2-7} \cmidrule(lr){8-11}
\textbf{} & 2014 & 2015 & 2016 & 2017 & \cellcolor{lighterblue}PTR & \cellcolor{lighterpink}{2018} & Full & Anon & \cellcolor{lighterpink}{Strip} & \cellcolor{lighterblue}XRep \\
\midrule
Ensemble Voting         & 0.827 & 0.860 & 0.904 & 0.846 & \cellcolor{lighterblue}\textbf{0.905} & \cellcolor{lighterpink}\textbf{0.876} & 22.60& 11.01& \cellcolor{lighterpink}8.39& \cellcolor{lighterblue}14.00\\
Ensemble Averaging      & 0.829 & \textbf{0.860} & 0.904 & 0.843 & \cellcolor{lighterblue}0.905 & \cellcolor{lighterpink}0.875 & 27.52& 14.83& \cellcolor{lighterpink}8.05& \cellcolor{lighterblue}16.80\\ \midrule
Weight Averaging~\cite{wortsman2022model}        & \textbf{0.859} & 0.857 & 0.891 & 0.777 & \cellcolor{lighterblue}0.891 & \cellcolor{lighterpink}0.773 & 25.06 & 14.17 & \cellcolor{lighterpink}\textbf{10.32} & \cellcolor{lighterblue}16.52 \\
Task Arithmetic~\cite{ilharco2022editing}         & 0.747 & 0.833 & 0.906 & \textbf{0.912} & \cellcolor{lighterblue}0.898 & \cellcolor{lighterpink}0.853 & \textbf{32.40} & \textbf{17.61} & \cellcolor{lighterpink}8.71 & \cellcolor{lighterblue}\textbf{19.57} \\
Ties-Merging~\cite{yadav2023ties}            & 0.791 & 0.835 & 0.897 & 0.853 & \cellcolor{lighterblue}0.889 & \cellcolor{lighterpink}0.875 & 32.38 & 16.55 & \cellcolor{lighterpink}7.84 & \cellcolor{lighterblue}18.92 \\
AdaMerging~\cite{yang2023adamerging}   & 0.751 & 0.839 & \textbf{0.908} & 0.909 & \cellcolor{lighterblue}0.895 & \cellcolor{lighterpink}0.852 & 24.77 & 14.01 & \cellcolor{lighterpink}6.62 & \cellcolor{lighterblue}15.13 \\ \midrule
\textbf{\sys{}~(Ours)}  & \textbf{0.867} & \textbf{0.854} & \textbf{0.894} & \textbf{0.890} & \cellcolor{lighterblue}\textbf{0.923} & \cellcolor{lighterpink}\textbf{0.940} & \textbf{36.97} & \textbf{25.42} & \cellcolor{lighterpink}\textbf{15.05} & \cellcolor{lighterblue}\textbf{25.81} \\
\bottomrule
\end{smalltabularx}%
\begin{tablenotes}[flushleft]
    \fontsize{6}{8}\selectfont
    \item[*] Rows 1–2 are ensemble-based baselines, while Rows 3–6 list model-merging approaches. For each metric (column), the best-performing value among the six baselines is highlighted in bold; results of our method are bolded throughout for ease of comparison. Some entries may appear numerically identical due to rounding, yet only the true maximum is highlighted.
    \item[*] Columns 2–7 present results for malware detection, and Columns 8–11 for binary analysis. Within each application, {\setlength{\fboxsep}{0pt}\fcolorbox{black}{lightpink}{\phantom{\rule{0.8em}{0.8em}}}} marks the most security-critical scenario (new threats in 2018 or highly stripped binaries), whereas {\setlength{\fboxsep}{0pt}\fcolorbox{black}{lightblue}{\phantom{\rule{0.8em}{0.8em}}}} indicates generalization performance (retention over earlier years or cross-representation across decompilation levels).
\end{tablenotes}
\end{threeparttable}
\label{tbl:cmp_merging}
\end{table*}


\subsection{Effectiveness on Representation Shift} \label{sec:eva_cmp_cl_rep}

We next compare \sys{} against CL baselines on representation shift using the binary summarization benchmark. 

\nobf{Evaluation Metrics}
We use the BLEU score~\cite{papineni2002bleu} as the primary metric to evaluate the similarity between generated and reference summaries.
Given a candidate summary, BLEU is defined as:
$
\text{BLEU} = \text{BP} \cdot \exp\left(\sum_{n=1}^{N} w_n \log p_n \right)
$,
where $p_n$ is the $n$-gram precision, $w_n$ is the corresponding weight (usually uniform, $1/N$), and $\text{BP}$ is the brevity penalty to discourage overly short outputs.
We report BLEU scores for all three representations to capture both local and cumulative learning effects.
Among them, the stripped binaries represent the most security-critical setting and serve as the primary target for continual adaptation.
To measure overall generalization across representations, we take the average to define a Cross-Representation (XRep) metric as:
$
\text{XRep}(R, f) = \frac{1}{R} \sum_{r=1}^{R} f(r)
$,
where $f(r)$ denotes a performance measure (e.g., BLEU, EM, METEOR~\cite{banerjee2005meteor}) evaluated on representation level $r$, and $R$ is the total number of representation levels.
This metric is a direct analogy to the AUT metric used in previous year-evolving temporal task; here, it measures the model's average performance across representations rather than time.
For completeness, we report complementary semantic similarity metrics \rcolor{in the supplementary open-source artifacts repository~(which show trends consistent with BLEU)} and include a small-scale manual analysis in \autoref{sec:summary_analysis}. 

\begin{figure}[t]
\vspace{-5pt}
    \centering
    \includegraphics[width=.92\linewidth]{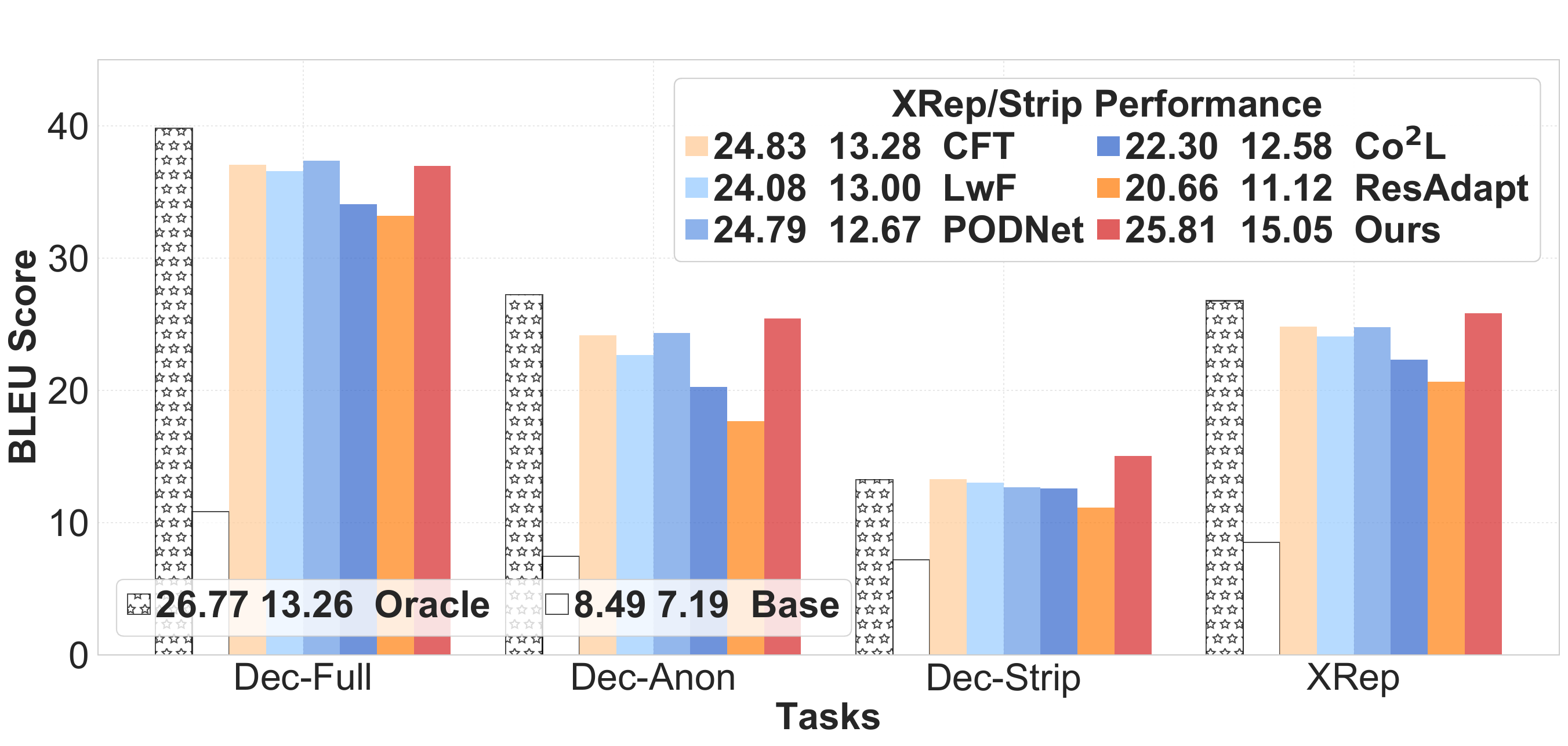}
    \vspace{-1em}
    \caption{Comparison with CL baselines in binary analysis.}
    \label{fig:cmp_cl_binary}
\end{figure}

\nobf{Result Analysis}
We summarize the performance of  all CL baselines and \sys{} in~\autoref{fig:cmp_cl_binary}, where models are sequentially trained across decompilation levels to evaluate cross-representation generalization~(XRep) and the security-critical stripped binaries~($\text{BLEU}_\text{strip}$).
Unlike the temporal drift setting, all other baselines underperform the naive CFT, with XRep scores lower by $0.04\sim4.17$ and $\text{BLEU}_\text{strip}$ lower by $0.28\sim2.16$.
The architecture-based ResAdapt exhibits the weakest performance on both metrics.
This degradation likely arises from its task-specific expansion strategy: while isolating parameters helps mitigate forgetting in disjoint tasks, it disrupts the cumulative transfer needed to generalize across related abstraction levels.
Among the distillation-based methods, PODNet attains the highest XRep, which is $0.71$ higher than the second-best LwF, but transfers knowledge less effectively to the most obfuscated binaries, dropping by~$0.33$ BLEU compared to LwF.
This trend suggests that existing CL methods, particularly those designed for incremental classification, are less effective in this scenario, as they fail to maintain consistent representation alignment.

In contrast, \sys{} achieves clear and consistent improvements on both metrics.
It raises the BLEU on stripped binaries to~$15.05$, substantially outperforming transfer-learning methods used in existing research~($7.20$, $+109.0\%$) and even exceeding the oracle~($13.26$, $+13.5\%$).
At the same time, \sys{} maintains strong performance across all decompilation levels, achieving an average XRep score of~$25.81$ and an improvement of~$3.9\%$ over the best baseline.
These results confirm that \rcolor{controlled} forgetting effectively stabilizes representation learning, enabling cumulative knowledge transfer from easier domains to more obfuscated, data-scarce representations without sacrificing overall generalization.


\subsection{Comparison to Aggregation Baselines} \label{sec:eva_cmp_ensemble}

We also compare \sys{} against alternative strategies that preserve individual models instead of sequential adaptation. 
These approaches aggregate model outputs~(ensemble) or parameters~(merging) after the training on all domains. 
We evaluate them across both applications, i.e., aggregating malware classifiers trained from different years and summarization models fine-tuned on distinct representation levels. 

\nobf{Evaluation Setup}
We adopt the previously established metrics for each application.
For malware detection, model-preserving methods are evaluated through a single aggregation step at the final year~(2018).
Each resulting model, including \sys{}, is then tested on all earlier years’ test sets as well as the final-year data.
Since these approaches do not involve sequential updates, AUT is not applicable, and PTR is instead computed by normalizing each year’s performance against the best score achieved by individually trained single-year models. 
For binary analysis, these methods are similarly adopted by fine-tuning an individual model from CodeT5 for each abstraction level. These models are then aggregated into a final model, evaluated across representation levels.

\nobf{Result Analysis}
We present the overall results in~\autoref{tbl:cmp_merging}.
In malware detection, we observe an interesting pattern in the model-preserving baselines. 
Although unit models are trained concurrently, the performance of their aggregated versions is not uniform across all tasks. 
Instead, they exhibit a distinct performance boost on the middle years (e.g., 2016 and 2017) compared to the edge years.
This fluctuation likely arises from partial overlap in malware families or shared behavioral patterns between adjacent years, which causes over-optimization for central tasks during model aggregation. 
Consequently, this non-uniformity leads to two critical failures. First, these baselines show a severe inability to detect new malware, where the best F1-score on 2018 data is only $0.876$, which \sys{} surpasses by $7.3\%$.
Second, their performance variance across earlier years is substantially higher, with standard deviations $1.7\sim3.9$ times greater than that of \sys{}, dragging their PTR scores $2.7\%$ lower on average.

Unlike the temporal drift scenario, the representation gaps between abstraction levels are more pronounced in binary analysis. 
As a result, model-merging approaches generally outperform ensemble-based ones, suggesting that direct parameter consolidation better mitigates discrepancies across code abstractions than output-level aggregation.
Nevertheless, \sys{} surpasses all model-preserving baselines by a substantial margin, improving BLEU by~$45.8\%$ on stripped binaries and~$31.9\%$ in XRep compared to the best-performing baselines, i.e., Weight Averaging and Task Arithmetic, respectively.
Compared to prior results in~\autoref{fig:cmp_cl_malware} and~\autoref{fig:cmp_cl_binary}, none of the aggregation baselines surpass existing CL methods, despite their `unfair' advantage of concurrent access to all task models. 
This contrast likely stems from the non-IID and highly evolving nature of security data, where static aggregation fails to reconcile domain shifts that violate the assumptions on which these methods rely~\cite{wu2023multi}. 
Overall, these results reaffirm a key distinction between preservation and accumulation: while model-preserving approaches retain fragmented expertise across tasks, \sys{} consolidates knowledge effectively through controlled continual adaptation.


\subsection{Ablation Analysis and Case Studies}

\subsubsection{Ablation Analysis}  \label{sec:eva_ablation}
\begin{figure}[t]
\vspace{-6pt}
    \centering 
    \subfloat[Weight Averaging\label{fig:sub1}]{%
        \includegraphics[width=0.49\linewidth]{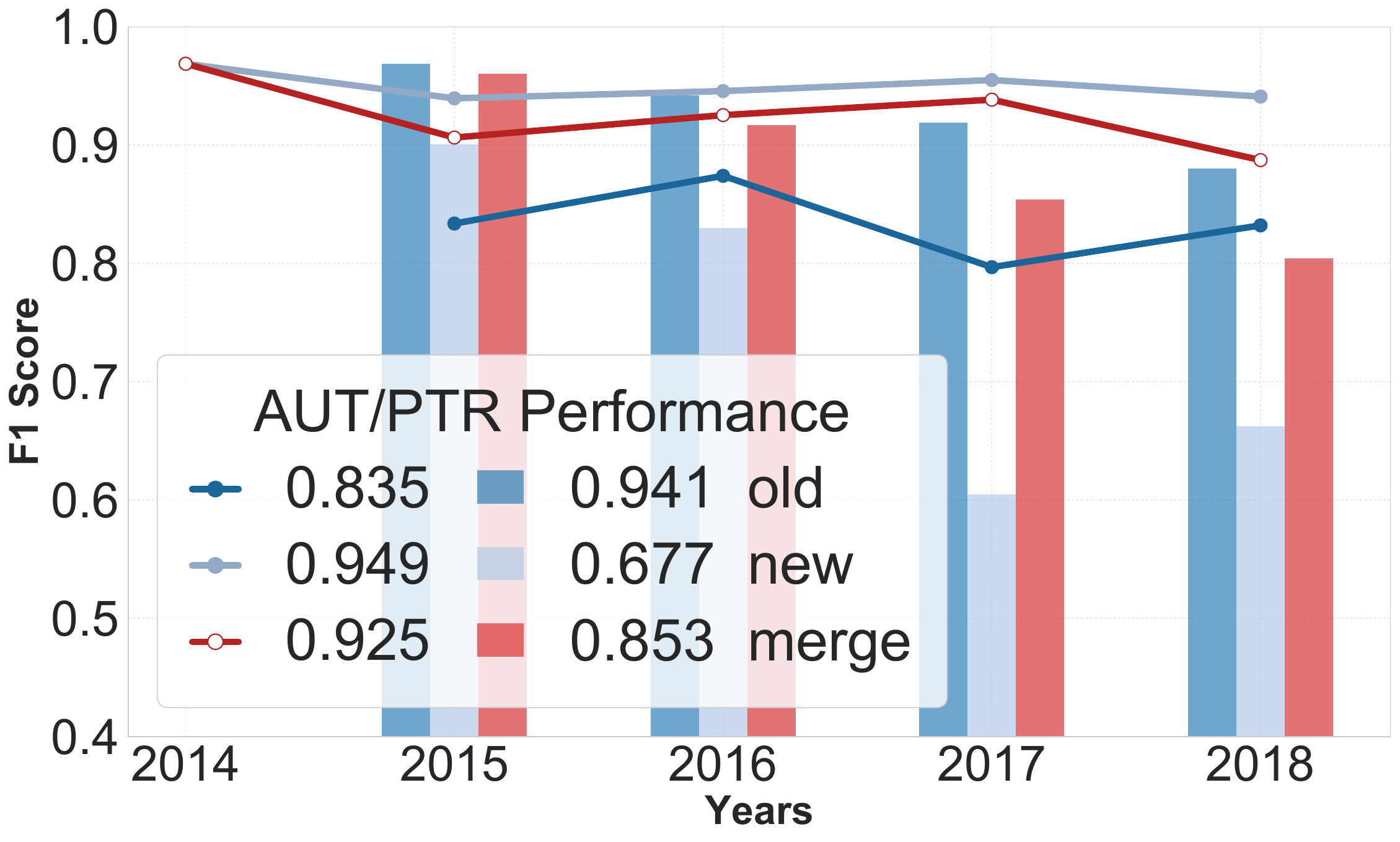}
    }
    \subfloat[Task Arithmetic\label{fig:sub4}]{%
        \includegraphics[width=0.49\linewidth]{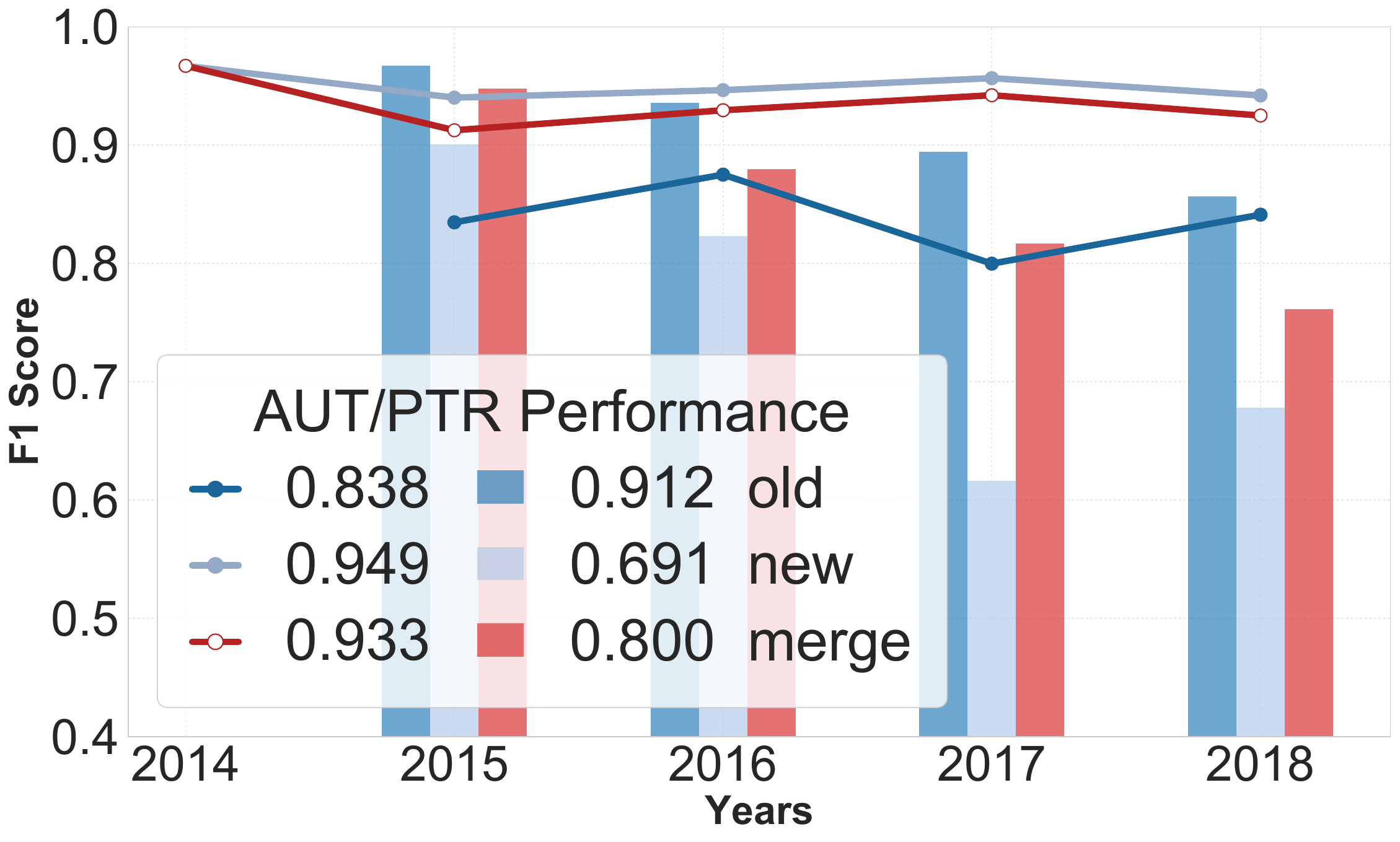}
    }
    \hfill 
    \subfloat[Ties-Merging\label{fig:sub2}]{%
        \includegraphics[width=0.49\linewidth]{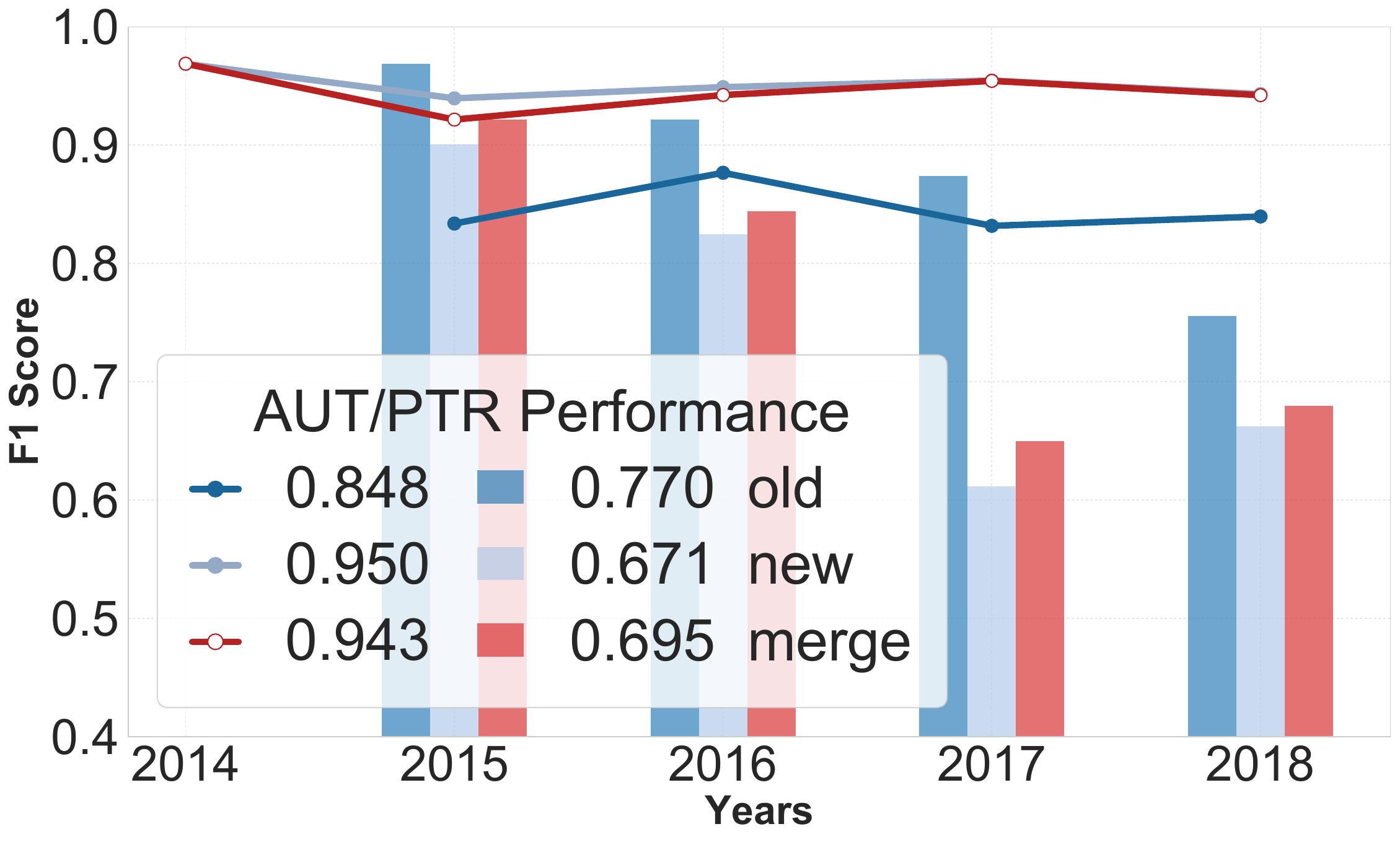}
    }
    \subfloat[AdaMerging\label{fig:sub5}]{%
        \includegraphics[width=0.49\linewidth]{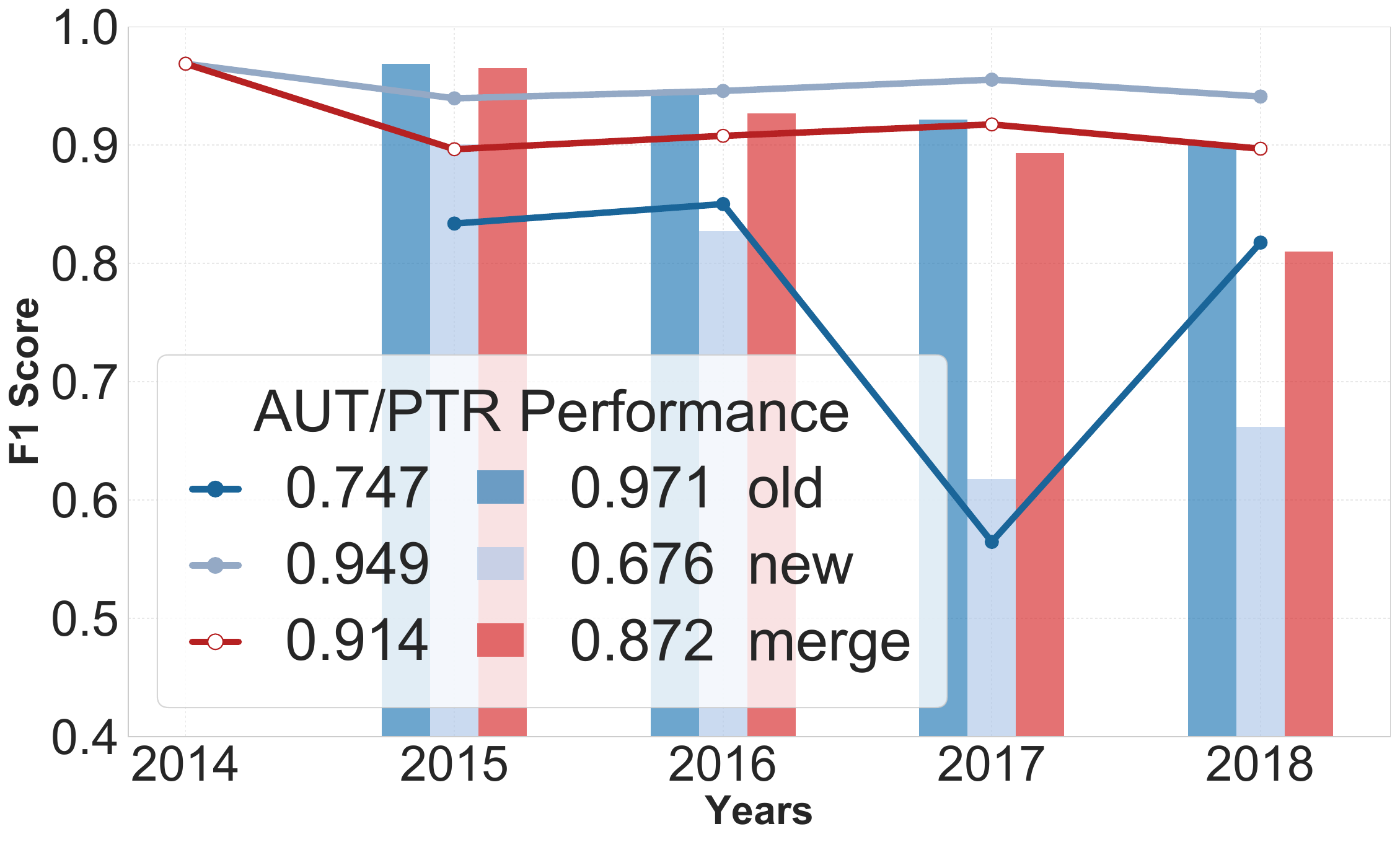}
    }
    \hfill 
    \subfloat[w/o Low-rank\label{fig:sub3}]{%
        \includegraphics[width=0.49\linewidth]{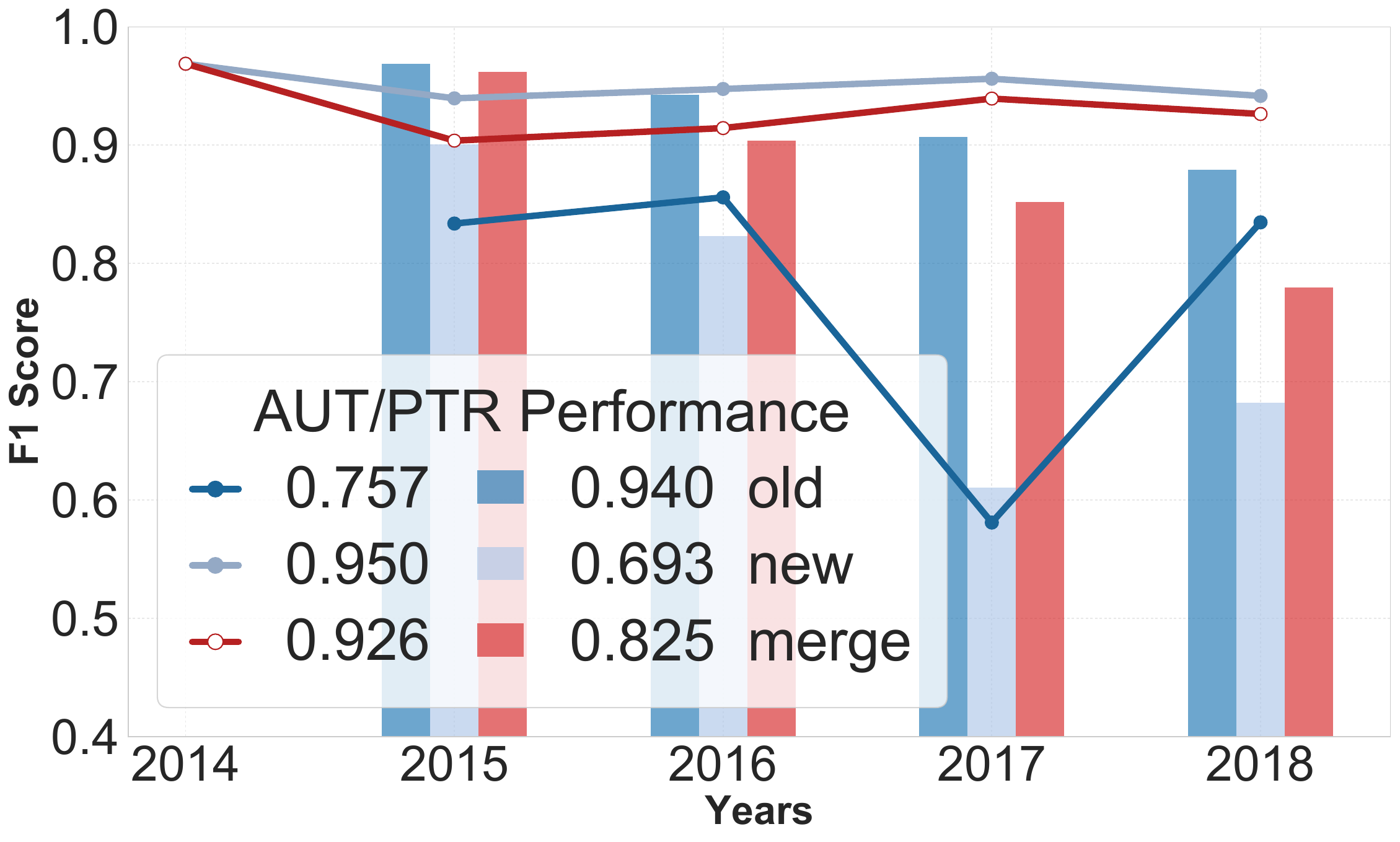}
    }
    \subfloat[\sys{}\label{fig:sub6}]{%
        \includegraphics[width=0.49\linewidth]{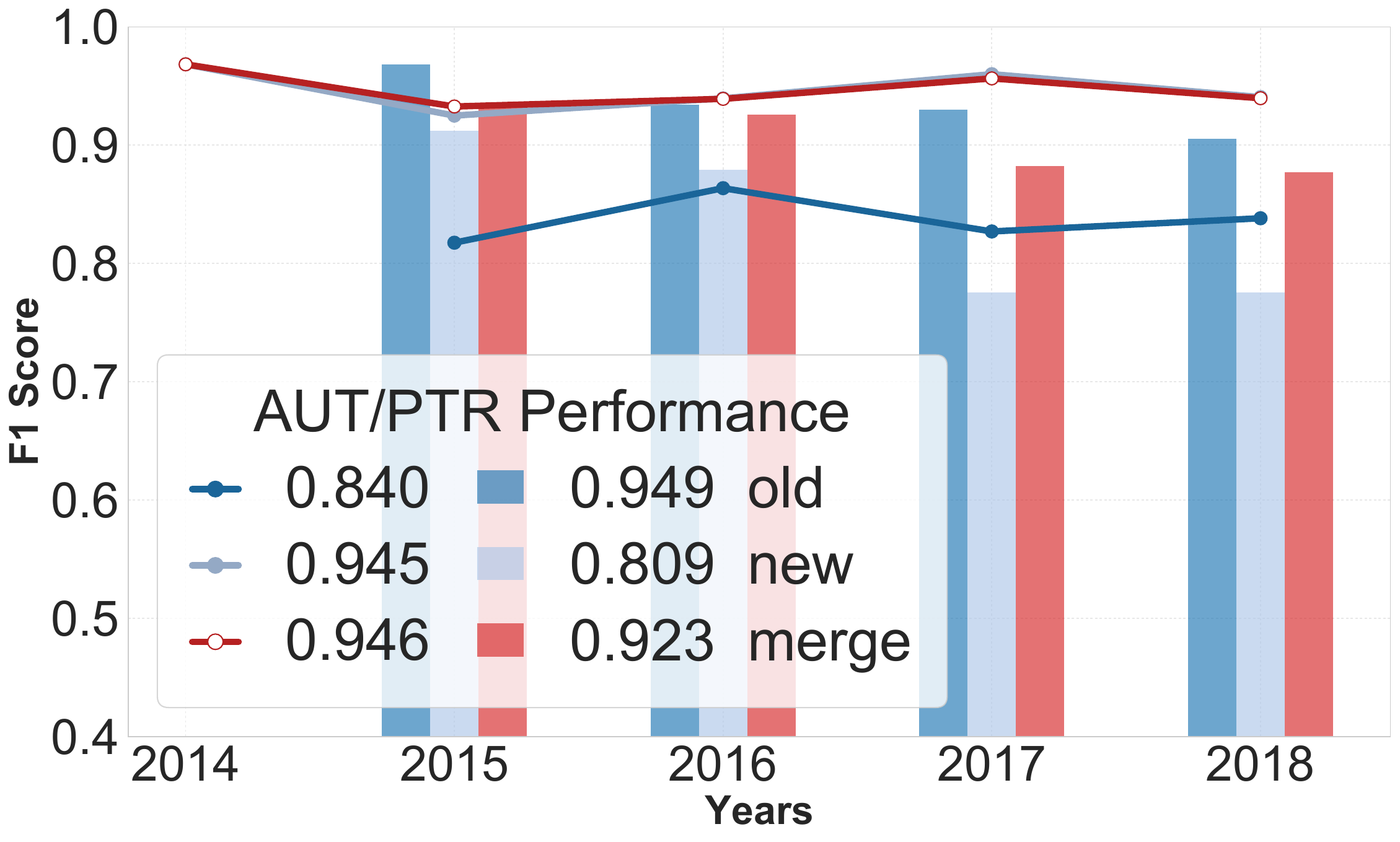}
    }
    \vspace{-5pt}
    \caption{Ablation analysis of \sys{}.}
    \label{fig:ablation_analysis}
\end{figure}

We conduct an ablation study on the malware detection task to validate the contributions of two main components in \sys{}, with results presented in \autoref{fig:ablation_analysis}.

\nobf{Adaptive Merging Strategy}
First, to validate the effectiveness of our adaptive merging strategy, we replace it with the four prominent model merging techniques, applying them in the same continual, data-free setting (i.e., merging the old model $f_{\text{old}}^{t=n} \leftarrow f_{\text{merge}}^{t=n-1}$ with the new model $f_{\text{new}}^{t=n}$). 
The adapted baseline results are presented in the first two rows, where we report the performance of both constituent models and their merged outcome at each year, allowing us to analyze how different merging strategies balance knowledge.
First, Weight Averaging and Task Arithmetic show moderate degradation in both AUT and PTR, confirming that simple linear fusion cannot simultaneously balance old and new knowledge.
TIES-Merging maintains adaptability~(AUT~$-0.32\%$) but its rigid parameter alignment causes it to closely mimic the new model, severely compromising retention~(PTR~$-24.70\%$).
AdaMerging achieves the best retention among baselines, showing that adaptive parameter scaling helps preserve shared representations.
However, its AUT decreases the most and its PTR remains~$5.53\%$ lower than \sys{}, indicating that \sys{}’s low-rank, confidence-guided consolidation yields more stable forgetting control.


\nobf{Low-Rank Constraint}
Second, we investigate the role of the low-rank structural constraint. We create a variant that removes the low-rank decomposition. In this setup, the new model is fine-tuned on the new data within the parameter space $\mathbb{R}^{d_{\text{in}}\times d_{\text{out}}}$, and our adaptive sparse mask is also learned in that space to merge it with the old model.
Results in the last row of the figure confirm that full parameter space merging violates our interference constraints.
It induces severe parameter drift, breaking the adaptive mask's ability to consolidate and causing a significant performance drop in both AUT ($-2.11\%$) and PTR ($-10.62\%$).
Even so, arbitration alone maintains balanced performance, ranking mid-tier among the four adapted baselines, where two surpass it in AUT and the other two do so in PTR.
Compared to AdaMerging, which is adaptive in learning layer-wise coefficients through entropy minimization, our arbitration employs confidence-guided consolidation that pulls logits toward new model when old model’s predictions are uncertain, thereby achieving higher AUT but lower PTR, a trade-off for protecting new, security-critical knowledge.

\begin{figure}[t]
\vspace{-8pt}
    \begin{minipage}{.56\linewidth}
    \includegraphics[width=\linewidth]{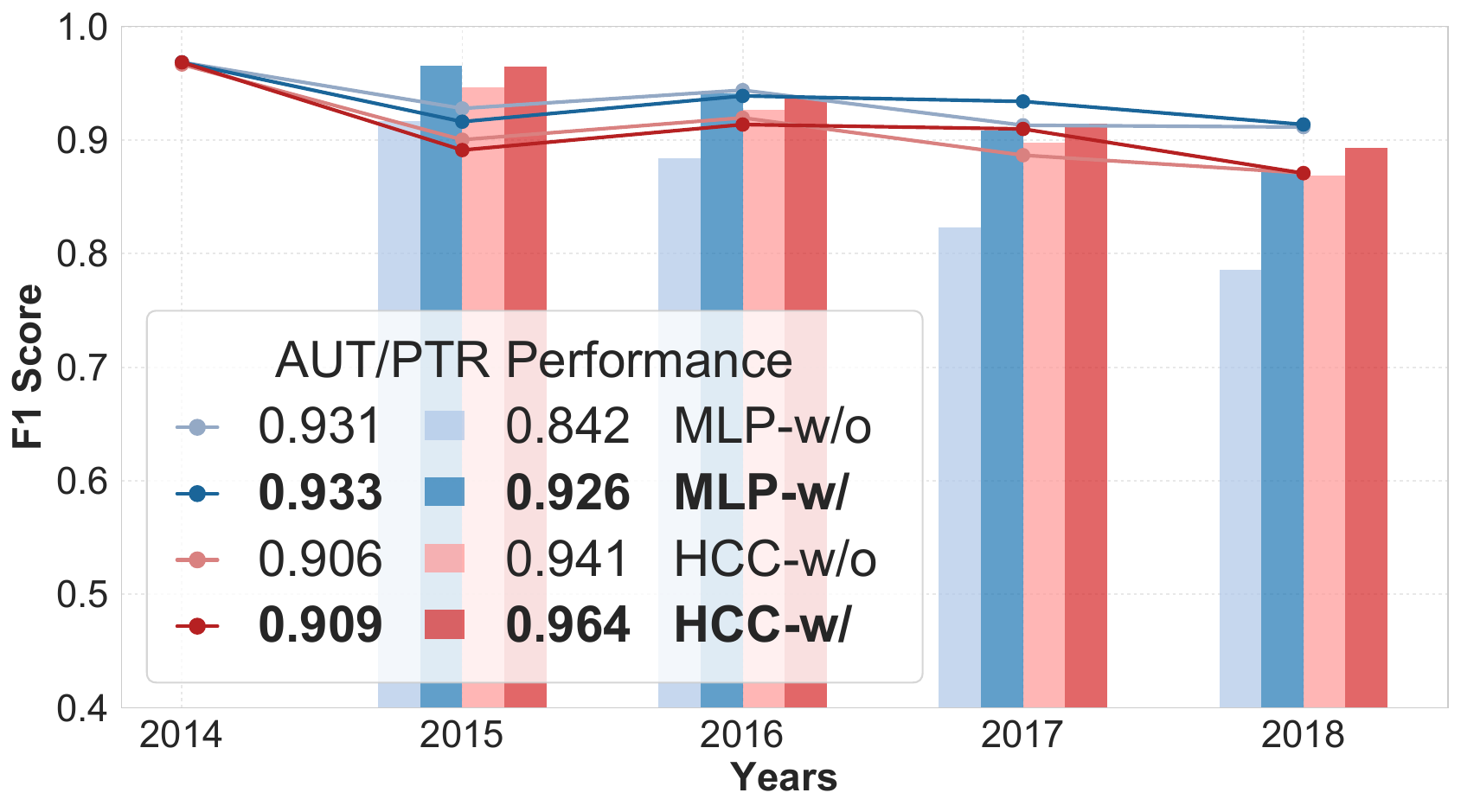}  
    \end{minipage}
    \begin{minipage}{0.41\linewidth}
    \centering
    \renewcommand{\arraystretch}{1.16} 
    \resizebox{\linewidth}{!}{
    \begin{tabular}[b]{cc|cc|cc|}
    \cline{3-6}
                                                &      & \multicolumn{2}{c|}{MLP}       & \multicolumn{2}{c|}{HCC}       \\ \cline{3-6} 
                                                &      & \multicolumn{1}{c|}{Old} & New & \multicolumn{1}{c|}{Old} & New \\ \hline
    \multicolumn{1}{|c|}{\multirow{3}{*}{2015}} & w/o  & \multicolumn{1}{c|}{0.917}    &  0.928   & \multicolumn{1}{c|}{0.947}    &   0.900  \\ \cline{2-6} 
    \multicolumn{1}{|c|}{}                      & w/   & \multicolumn{1}{c|}{0.966}    &  0.916   & \multicolumn{1}{c|}{0.965}    &   0.891  \\ \cline{2-6} 
    \multicolumn{1}{|c|}{}                      & $\Delta$ & \multicolumn{1}{c|}{\impup{5.33}}    &   \impdown{1.26}  & \multicolumn{1}{c|}{\impup{1.92}}    &   \impdown{1.01}  \\ \hline
    \multicolumn{1}{|c|}{\multirow{3}{*}{2016}} & w/o  & \multicolumn{1}{c|}{0.884}    &   0.944  & \multicolumn{1}{c|}{0.926}    &   0.920  \\ \cline{2-6} 
    \multicolumn{1}{|c|}{}                      & w/   & \multicolumn{1}{c|}{0.941}    &  0.939   & \multicolumn{1}{c|}{0.938}    &   0.914  \\ \cline{2-6} 
    \multicolumn{1}{|c|}{}                      & $\Delta$ & \multicolumn{1}{c|}{\impup{6.45}}    &   \impdown{0.54}  & \multicolumn{1}{c|}{\impup{1.20}}    &  \impdown{0.64}   \\ \hline
    \multicolumn{1}{|c|}{\multirow{3}{*}{2017}} & w/o  & \multicolumn{1}{c|}{0.823}    &   0.913  & \multicolumn{1}{c|}{0.897}    &   0.887  \\ \cline{2-6} 
    \multicolumn{1}{|c|}{}                      & w/   & \multicolumn{1}{c|}{0.908}    &   0.934  & \multicolumn{1}{c|}{0.915}    &   0.910  \\ \cline{2-6} 
    \multicolumn{1}{|c|}{}                      & $\Delta$ & \multicolumn{1}{c|}{\impup{10.3}}    &  \impup{2.29}   & \multicolumn{1}{c|}{\impup{1.91}}    &    \impup{2.61} \\ \hline
    \multicolumn{1}{|c|}{\multirow{3}{*}{2018}} & w/o  & \multicolumn{1}{c|}{0.786}    &   0.911  & \multicolumn{1}{c|}{0.869}    &   0.871  \\ \cline{2-6} 
    \multicolumn{1}{|c|}{}                      & w/   & \multicolumn{1}{c|}{0.872}    &   0.914  & \multicolumn{1}{c|}{0.893}    &  0.871   \\ \cline{2-6} 
    \multicolumn{1}{|c|}{}                      & $\Delta$ & \multicolumn{1}{c|}{\impup{11.1}}    &   \impup{0.25}  & \multicolumn{1}{c|}{\impup{2.84}}    &     \impnone\\ \hline
    \end{tabular}
    }
    \end{minipage}   
    \caption{Improvements in retention \& adaptation when combining \sys{} with active learning for different classifiers.}
    \label{tbl:combine_hcc}
    \vspace{-5pt}
\end{figure}

\subsubsection{Combination with Active Learning}  \label{sec:eva_combine_hcc} 

In malware detection, recent research adopts \emph{active learning} to support model adaptation under limited human-labeling budgets.
These approaches typically focus on detecting and labeling new drift samples, using them for retraining or fine-tuning, during which most methods still rely on CFT.
Among them, HCC~\cite{continuous} represents a state-of-the-art framework that designs a classifier with hierarchical contrastive learning to intrinsically detect drift during deployment.
To evaluate our compatibility with such paradigms, we integrate \sys{} into HCC’s best-performing warm-start procedure.
In each iteration, informative samples are selectively labeled under a fixed budget of $5000$, after which model adaptation is performed using \sys{} to assess the improvements over the original adaptation scheme.
For completeness, we also include the standard MLP classifier used in our previous experiments, placing it under the same active learning protocol~(where drift samples are selected by HCC's pseudo loss sampler) and adapting it with \sys{} for direct comparison.

\nobf{Result Analysis}
As indicated in~\autoref{tbl:combine_hcc}, the integration of \sys{} into active learning consistently improves both classifiers.
For the MLP classifier, \sys{} increases AUT by~$0.2\%$ and PTR by~$10.0\%$, while for the more advanced HCC, it yields additional gains of~$0.3\%$ in AUT and~$2.4\%$ in PTR.
Higher retention is observed for the MLP classifier compared to the earlier continual fine-tuning results (PTR $0.842$ vs.~$0.666$), although adaptability slightly declines due to the limited labeling budget (AUT $-2.0\%$). 
\sys{} restores its adaptability and further strengthens retention, confirming that preservation of our core advantages even when the data-access assumptions differ from retrospective-free setting.
The HCC classifier, benefiting from hierarchical contrastive representations, remains the most robust baseline across years, achieving a strong retention score~(PTR $0.941$) but at the expense of adaptability (AUT lower than MLP by~$2.7\%$).
Nevertheless, incorporating \sys{} raises both metrics simultaneously.
A closer year-wise analysis reveals that \sys{} consistently improves performance on old tasks, particularly in later years, where accumulated knowledge becomes more critical.
While minor fluctuations occur in early-year new-test results, the strengthened forward transfer achieved through \rcolor{controlled} forgetting ultimately yields superior long-term performance.





\subsubsection{Summarization Quality Analysis} \label{sec:summary_analysis}

While automated metrics like BLEU provide quantitative results,
we conduct a qualitative manual analysis, with two primary goals:
1)~to understand why the automated scores for stripped binaries are universally low; 2)~to demonstrate how \sys{} provides more interpretable and useful summaries. 

%





\begin{figure}[t]
    \centering
    \includegraphics[width=\linewidth]{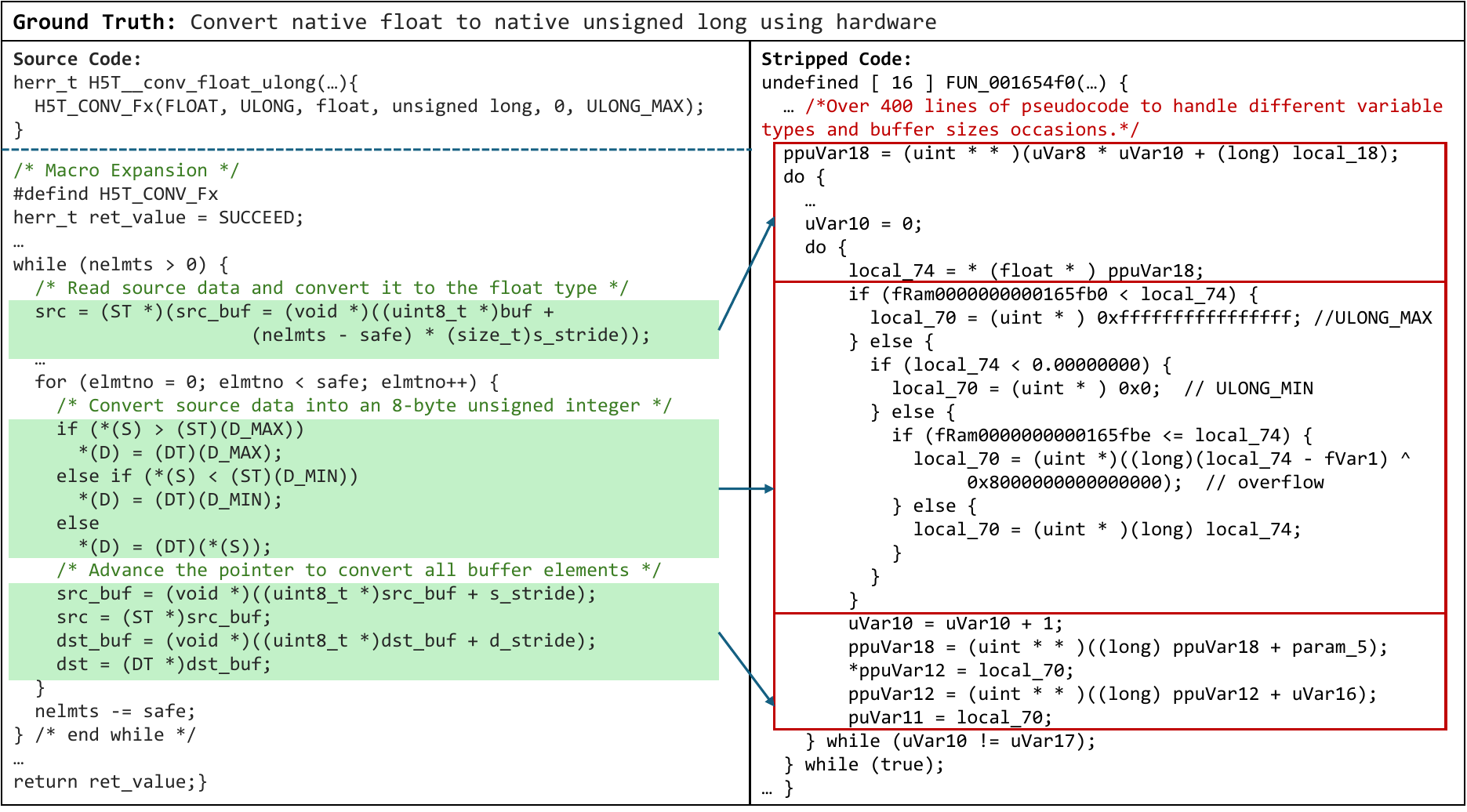}
    \caption{An example of how the function decompilation from a stripped binary significantly expands the source code. \sys{} achieves an exact match with BLEU of $100$, while baselines fail (CFT: $4.92$, Oracle: $20.16$).}
    \vspace{-5pt}
    \label{fig:case_study_1}
\end{figure}

\begin{figure}[t]
    \centering
    \includegraphics[width=\linewidth]{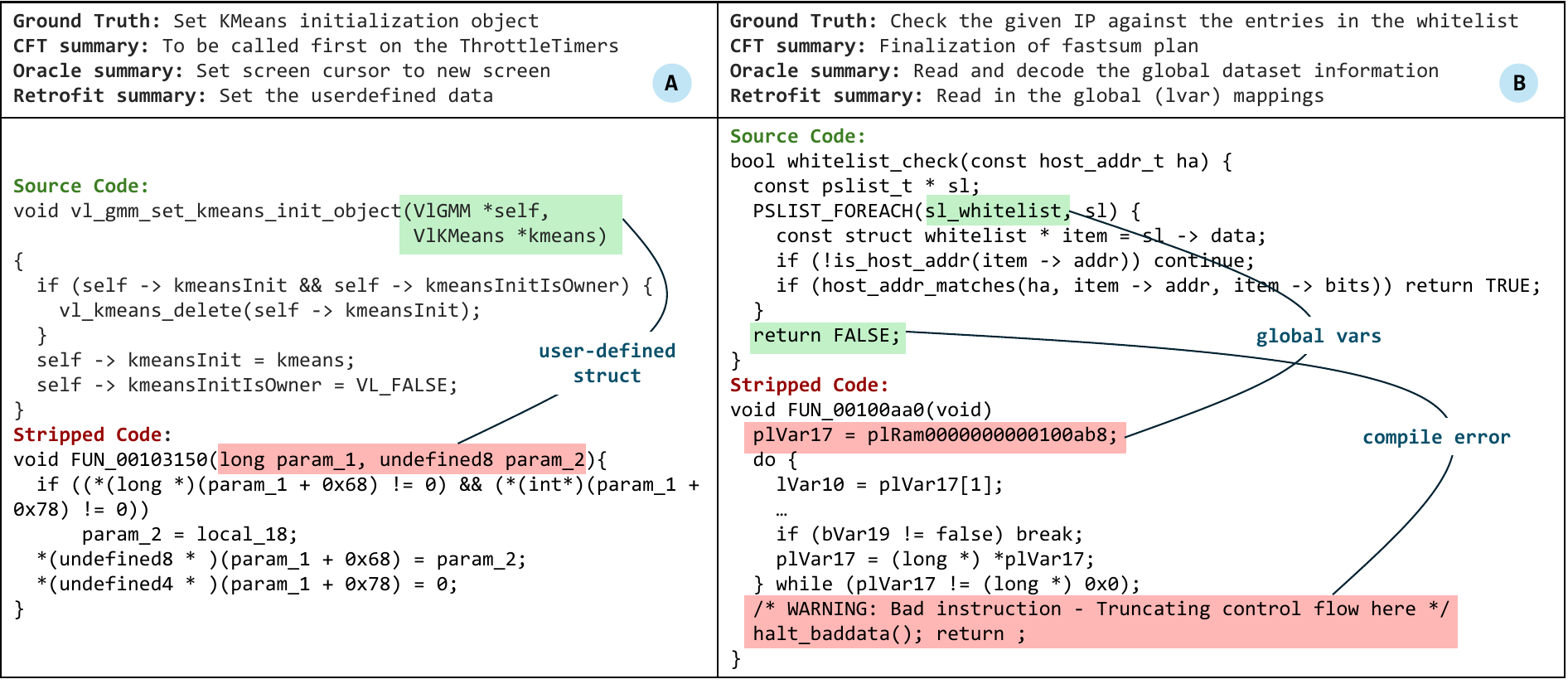}
    \caption{Summary comparison for challenging stripped binaries, where the BLEU of all methods is lower than $20$.}
    \vspace{-5pt}
    \label{fig:case_study_2}
\end{figure}

\nobf{Exact Match on Complex Logic}
We provide an example in~\autoref{fig:case_study_1} to show the difficulty of summarizing stripped binaries.
The original source function~(left) already contains intricate logical operations, while its stripped counterpart~(right) further obscures semantic intent. 
Compiler preprocessing and optimizations, such as macro expansion and loop unrolling, inflate a single source statement into more than $400$ lines of decompiled pseudocode. 
\sys{} achieves a BLEU of~$100$ in this example, exactly matching the ground truth.
It correctly identifies both the operation (`float to unsigned long' conversion) and its implementation detail (`using hardware'), inferred from the three highlighted code patterns.
By contrast, the best CL baseline only reaches $4.92$ BLEU, misidentifying the source type and the operational detail.
The oracle model also performs poorly, summarizing the function as `Convert native unsigned integer to native float...', inverting the conversion and failing to capture the correct semantics.
This case underscores \sys{}’s superior capability to recover precise functional intent from heavily obfuscated binaries.

\nobf{Semantic Robustness on Limited information}
\autoref{fig:case_study_2} presents two challenging cases where all methods achieve BLEU scores below~20. 
We identify two factors that explain the universally low scores.
    1)~\emph{Abstract ground truth}: reference summaries often describe high-level intent unavailable in stripped code, e.g., the metadata of parameter \texttt{VlKMeans} in Case~A and variable \texttt{sl\_whitelist} in Case~B; 
    2)~\emph{Decompilation artifacts}: the decompiler itself introduces errors, e.g., in Case B, the decompiler fails to propagate a boolean return value, causing truncated control flow. 
Despite the low scores, \sys{} produces the most faithful interpretation of the observable logic, whereas baselines often produce irrelevant or hallucinated descriptions.
In Case A, its summary accurately captures the data initialization behavior, while both CFT and Oracle hallucinate semantics, misidentifying \texttt{param\_1} as unrelated entities (e.g., a \emph{ThrottleTimer} or mouse coordinates).
In Case B, \sys{} likewise offers the most plausible interpretation, correctly recognizing the global-variable access pattern in the partial control flow.
By contrast, CFT fabricates irrelevant logic, seemingly driven by superficial lexical cues, whereas Oracle overgeneralizes, mistaking conditional whitelist checking for dataset processing.

\subsection{\rcolor{Computational and Robustness Analysis}} \label{sec:compute_robust}

\rcolor{We next analyze RETROFIT’s computational efficiency and practical robustness under varying update capacity~(e.g., rank and sparsity) and task horizon, supported by empirical statistics from the malware detection application.}

\nobf{Computational Cost}
\rcolor{
We evaluate the practical overhead of \sys{} by comparing its per-epoch training time and GPU memory usage with CFT and ResAdapt.
CFT represents the most efficient baseline, while ResAdapt represents architecture-based CL methods with network expansion.
As shown in~\autoref{tab:time_memory}, \sys{} achieves the lowest GPU memory usage.
Its per-epoch runtime is slightly higher than CFT, but remains substantially lower than ResAdapt.
}
\rcolor{Unlike architecture-based methods,} \sys{} maintains a constant model size and preserves the same single forward path as the base model at inference.
Compared with the simplest CFT baseline, which trains all parameters at a cost of $O(d_{\text{in}}d_{\text{out}})$ per task, \sys{} reduces the training complexity by approximately a factor of $r/\min(d_{\text{in}},d_{\text{out}})$.
\rcolor{This primarily lowers update-related memory, including gradients and optimizer states.
The modest runtime overhead over CFT mainly comes from mask optimization, whose cost scales linearly with the rank $r$. Arbitration adds only negligible overhead, i.e., requiring two additional forward passes through the teacher models.}
 
\begin{table}[t]
\centering
\captionsetup{labelfont={color=tcolor}, textfont={color=tcolor}}
\caption{Training time and GPU memory usage comparison. 
}
\vspace{-8pt}
\label{tab:time_memory}
\rcolor{
\begin{smalltabularx}{.92\linewidth}{l|C|C}
\toprule
\textbf{Method} & \textbf{Time (s/epoch)} & \textbf{Max alloc / reserved (MB)} \\
\midrule
CFT       & \cellcolor{lightergreen}$1.07 \pm 0.01$ & $181.13$ / $226.00$ \\
ResAdapt  & $1.67 \pm 0.09$ & $316.20$ / $350.00$ \\
\sys{}  & $1.23 \pm 0.03$ & \cellcolor{lightergreen}$150.19$ / $170.00$ \\
\bottomrule
\end{smalltabularx}
}
\end{table}


\nobf{\rcolor{Hyperparameter Sensitivity}}
\rcolor{
The main capacity hyperparameter of \sys{} is the low-rank dimension $r$, which controls the size of the update subspace; another important hyperparameter is the sparsity-balancing coefficient $\mu$, which shapes how concentrated the merged update becomes.
As shown in~\autoref{tab:sensitivity}, \sys{} remains stable across a broad range of both parameters.
We do not expect monotonic behavior in $r$: smaller ranks reduce potential interference, but can also limit the capacity needed to learn transferable updates; larger ranks provide more plasticity, while masking and arbitration control interference.
Varying $\mu$ introduces only minor performance changes, suggesting that it mainly affects the convergence path toward an effective sparsity pattern rather than the final stability and plasticity trade-off.
In practice, we recommend choosing a moderate rank to balance adaptation capacity and interference control, and setting $\mu$ within a stable range rather than tuning it aggressively.
}

\begin{table}[t]
\vspace{-5pt}
\centering
\captionsetup{labelfont={color=tcolor}, textfont={color=tcolor}}
\caption{Sensitivity analysis of $r$ and $\mu$.
The reference setting ($r=512$, $\mu=2\times10^{-6}$) is used in the main experiments.}
\vspace{-8pt}
\label{tab:sensitivity}
\rcolor{
\begin{smalltabularx}{\linewidth}{l|C|CCC|CCC}
\toprule
\multirow{2}{*}{\textbf{Metric}}
& \multirow{2}{*}{\textbf{Ref.}}
& \multicolumn{3}{c|}{\textbf{Varying $\bm r$}}
& \multicolumn{3}{c}{\textbf{Varying $\bm \mu$}} \\
\cmidrule(lr){3-5}\cmidrule(lr){6-8}
&
& $256$ & $768$ & $1024$
& $10^{-6}$ & $4{\times}10^{-6}$ & $8{\times}10^{-6}$ \\
\midrule
AUT
& \cellcolor{lightergreen}$0.946$
& $0.943$ & \cellcolor{lighterblue}$0.947$ & $0.944$
& $0.944$ & \cellcolor{lighterblue}$0.945$ & \cellcolor{lighterblue}$0.945$ \\
PTR
& \cellcolor{lightergreen}$0.923$
& $0.864$ & $0.923$ & \cellcolor{lighterblue}$0.928$
& \cellcolor{lighterblue}$0.921$ & $0.920$ & $0.918$ \\
\bottomrule
\end{smalltabularx}
}
\end{table}

\begin{figure}[t]
    \centering
    \includegraphics[width=\linewidth]{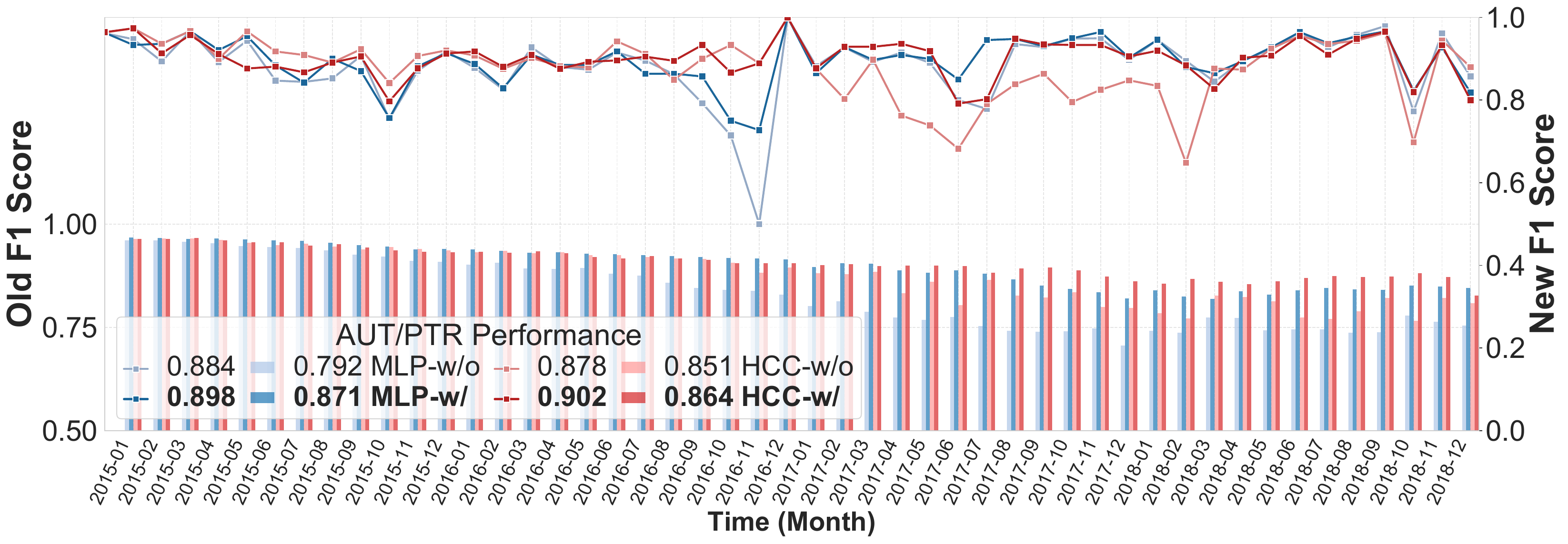}
    \vspace{-1em}
    \captionsetup{labelfont={color=tcolor}, textfont={color=tcolor}}
    \vspace{-5pt}
    \caption{Month-wise active learning-based malware adaptation w/ and w/o \sys{}, where the task number is 48.}
    \label{fig:hcc_monthly}
    \vspace{-1em}
\end{figure}

\nobf{\rcolor{Robustness to Update Frequency}}
\rcolor{
To study scalability as the number of tasks increases, we evaluate a month-wise active-learning setting with 48 sequential updates and a labeling budget of $500$ samples per month, complementing the year-wise setting in~\autoref{sec:eva_combine_hcc}.
As shown in~\autoref{fig:hcc_monthly}, \sys{} consistently improves both AUT and PTR across two models.
For MLP, \sys{} improves AUT by $1.6\%$ and PTR by $10.0\%$; for HCC, it improves AUT by $2.7\%$ and PTR by $1.5\%$.
Notably, finer-grained update cycles with less tuning data amplify interference, underscoring the need for our explicit control.
Compared with the year-wise setting, the relative AUT gains are larger on both models, while the PTR gains remain similar, suggesting that \sys{} scales to longer horizons.
}

\section{Discussion} \label{sec:discussion} 

\nobf{\rcolor{Oracle Upper Bound}}
\rcolor{Oracle remains the strongest overall reference, but access to all accumulated data does not eliminate interference. Under temporal or representation shift, joint training must fit heterogeneous past and current distributions with a single uniform objective, which can favor intermediate distributions while diluting specialization to the newest one. 
For example, this pattern is visible in~\autoref{tbl:cmp_cl_malware}, where Oracle performs better on new test data in the two middle years. Prior work has similarly observed that full training is not always the upper bound~\cite{wu2023multi}. Naive CFT can exceed Oracle on new data, but only with severe forgetting. Thus, \sys{}'s advantage on the new distribution is better understood as a regularization effect: it preserves the specialization benefit of sequential adaptation while controlling forgetting.}

\nobf{Applicable Domain}
As shown in~\autoref{sec:eva_combine_hcc}, \sys{} is compatible with different model architectures for malware detection. Likewise, it can be extended to emerging pretrained feature extractors like DexBERT~\cite{sun2023dexbert}.
In binary analysis, similar arbitration mechanisms can be applied to other LLMs~\cite{roziere2023codellama}.
Future work may further explore \sys{} under other forms of representation shift arising in security-critical contexts, e.g., binaries compiled with varying optimization levels, architectures, or platforms, and extend it naturally to related downstream tasks including binary similarity detection~\cite{ding2019asm2vec} and vulnerability discovery~\cite{liu2024exploring}.
More broadly, \sys{} offers a general solution for continual model adaptation in data-sensitive security applications. 

\nobf{Combining with Replay-based CL}  
While \sys{} is designed for data-restricted environments, it can be complemented with replay-based CL strategies when data retention policies are more permissive.    
In such settings, a small memory buffer provides representative samples from past tasks, improving coverage of underrepresented or highly heterogeneous behaviors.  
For instance, MADAR~\cite{rahman2025madar} demonstrates that distribution-aware replay enhances malware classification by accounting for heterogeneity both across and within families.  
Similar replay mechanisms can be adapted for different applications, and incorporating them into \sys{} has the potential to strengthen retention, but at the cost of additional storage and training overhead.

\section{Related Work} \label{sec:related}

\nobf{Representation Learning for Robustness}
Representation learning methods such as contrastive learning~\cite{chen2020simple, continuous} and invariant risk minimization~\cite{ahuja2020invariant} aim to extract invariant features that remain stable under distributional shifts. 
For example, hierarchical contrastive learning employed in HCC~\cite{continuous} enhances malware detection by separating benign and malicious samples while capturing intra-family semantics.
In security analytics, related efforts refine API representations through knowledge graphs~\cite{zhang2020enhancing} and structure-aware encoders that integrate control-flow information~\cite{pei2024exploiting}.
While sharing the goal of preserving performance amid changing conditions, they operate in static settings; instead, CL adapts representations dynamically as new datasets become available, making the two approaches naturally complementary.

\nobf{MTL and Model Merging}
Multi-task learning~(MTL)~\cite{zhang2021survey} learns shared representations across related tasks but requires simultaneous access to all task data. 
To bypass this constraint, model merging has emerged as an alternative to execute MTL, combining independently trained models post hoc rather than through joint training on all tasks~\cite{misra2016cross, wu2023pi}.
To mitigate task interference, TIES-Merging~\cite{yadav2023ties} applies a majority vote on the sign for each parameter, whereas AdaMerging~\cite{yang2023adamerging} learns adaptive merge weights at the task or layer level.
We integrate merging into CL settings and employ low-rank and adaptive sparse mask learning to regulate conflicts between old and new models.
Our ablation study in~\autoref{sec:eva_ablation} further confirms reduced interference when replacing our design with existing merging methods.

\nobf{Drift Detection and Active Learning}
Drift detection aims to identify samples that deviate significantly from the training distribution, often by monitoring prediction uncertainty or changes in the embedding distribution over time~\cite{lu2018learning, barbero2022transcending}.
In security analytics, such mechanisms have been widely used to detect emerging malware families or new attack behaviors~\cite{he2024combating, yang2021cade}.
Active learning complements this process by selecting those detected samples for human labeling, thus reducing annotation costs while guiding model retraining with an expanded dataset~\cite{zhang2020enhancing, continuous}.
As demonstrated in~\autoref{sec:eva_combine_hcc}, our method serves as an adaptation extension that seamlessly integrates into this online workflow. 

\nobf{Code Analysis for Security}
Recent advances in LLMs have opened new directions for automated code understanding in security applications. 
A key research question is how well these models comprehend low-readability code~\cite{llmcode24usenix, jin2023binary}. 
This exploration is often hampered by the scarcity of ground-truth data, which forces reliance on proxy metrics~\cite{he2025benchmarking}. 
Our work applies \sys{} to binary summarization, using a recent dataset compiled from open-source~(non-malware) projects that provides ground-truth summaries from source code comments~\cite{al2023extending}.
While this benchmark is valuable, we also note its remaining flaws~(discussed in \autoref{sec:summary_analysis}), highlighting the need for further dataset refinement in binary analysis. 
 

\section{Conclusion}
\label{sec:conclusion}

This paper tackles continual learning in data-sensitive binary security domains, where models must evolve without access to prior data.  
We identified catastrophic forgetting and distributional generalization failure as key barriers to long-term adaptability under such constraints.  
To address these, we propose \sys{}, which formulates knowledge accumulation as model consolidation and employs low-rank structural constraints with confidence-guided sparse regulation to achieve \rcolor{controlled} forgetting.
Empirical results across two representative applications, i.e., malware detection and binary analysis, demonstrate that \sys{} sustains strong retention and effective adaptation across model architectures and data distributions.  
These results establish \sys{} as a practical and effective foundation for trustworthy continual learning. 

\section*{Acknowledgments}
We would like to thank the anonymous reviewers and our shepherd for their insightful comments and valuable guidance to help improve this paper. This work is supported in part by Google ASPIRE, Google Academic Research Awards, the ``Pioneer'' and ``Leading Goose'' R\&D Program of Zhejiang (Grant No. 2024C01169), and the National Natural Science Foundation of China under Grants 62441238 and U2441240. 

\appendix
\section*{Ethical Considerations}

This work focuses on continual learning for defensive binary security applications, including malware detection and binary analysis. \rcolor{The main stakeholders are \emph{security analysts}, \emph{incident-response teams}, \emph{software vendors}, and \emph{end users}, all of whom may benefit from more robust model adaptation under distribution shift. By reducing adaptation failures as threats evolve or program representations degrade, such systems can improve detection reliability, analysis quality, and operational continuity.}

\rcolor{The main operational harms arise from misclassifications. \emph{False negatives} may allow malicious binaries to evade detection, delaying triage, remediation, or vulnerability discovery, harming end users and reducing trust in software vendors and security providers.
\emph{False positives} may generate unnecessary alerts, waste analyst effort, disrupt vendor release workflows, or block legitimate software for end users. Publication also carries limited \emph{dual-use risk}, since advances in binary analysis may be studied by adversarial users.}

This work is explicitly scoped to \emph{defensive use} and does not introduce new capabilities for malware creation, obfuscation, or evasion. 
\sys{} is intended to \emph{reduce adaptation failures under drift}, but not to replace human judgment in high-stakes settings. 
In addition, the retrospective-free design avoids storing or replaying historical binaries, which \emph{reduces dependence on long-term data access} to sensitive malware samples and proprietary software artifacts.

\rcolor{We judged that the expected benefits of improving robustness in defensive malware detection and binary analysis outweigh the remaining risks, particularly because the work can directly benefit stakeholders through improved binary analysis reliability while reducing dependence on sensitive data.}

\section*{Open Science}
We make the artifacts available at \href{https://zenodo.org/records/20339141}{Zenodo} and \href{https://github.com/XdlSecZJU/RETROFIT-CL}{GitHub}.
The shared artifacts consist of a complete implementation of our proposed continual learning framework \sys{}, along with all reproducible workflows necessary to adapt, merge, and evaluate models in the two applications described in the paper.  
A detailed README describes the code structure and usage. 
Due to licensing and redistribution constraints, the two datasets and pretrained BinT5 models are not included in the repository; instead, we provide instructions for obtaining them from their original sources and release our processed APK features and fine-tuned BinT5 checkpoints.

\bibliographystyle{plainurl}  
\bibliography{reference}      

\appendix

\section{Proofs} \label{appendix:proofs}

We consider a high-dimensional regime where the input dimension $d_{\mathrm{in}}$ is large relative to the update rank $r$. For any task $t$, the random projection matrix $A_t \in \mathbb{R}^{d_{\mathrm{in}} \times r}$ is initialized with i.i.d. entries such that $\mathbb{E}[(A_t)_{ik}] = 0$ and $\mathrm{Var}((A_t)_{ik}) = \sigma^2 > 0$ (as determined by the chosen initializer, e.g., Gaussian or Kaiming uniform).
In particular, a common normalization for random projections sets $\sigma^2 = 1/d_{\mathrm{in}}$.
Consequently, the following moment identities hold:
\begin{equation}\label{eq:moments_formal}\mathbb{E}[A_t^\top A_t] = \sigma^2 d_{\mathrm{in}} I_r, \qquad \mathbb{E}[A_t A_t^\top] = \sigma^2 r I_{d_{\mathrm{in}}}.\end{equation}
Given sub-Gaussian entries of $A_t$, for any fixed matrix $B \in \mathbb{R}^{r \times d_{\mathrm{out}}}$, the norm of $A_t B$ concentrates: 
for any $\epsilon\in(0,1)$,
\begin{equation}\label{eq:concentration_formal} \|A_t B\|_F^2 \ge (1-\epsilon) \sigma^2 d_{\mathrm{in}} \|B\|_F^2 \end{equation}
holds with probability at least $1-\delta$, where the failure probability satisfies $\delta \le 2 \exp\left(-c\,\epsilon^2 d_{\mathrm{in}}\right)$. 

\subsection{Proofs of Proposition 1} 

Let $s < t$ index an old task and a new task, respectively, and let $g_s := \nabla_\theta \mathcal{L}_s(\theta_{\mathrm{prev}}) \in \mathbb{R}^{d_{\mathrm{in}}\times d_{\mathrm{out}}}$ denote the gradient of a past task. 
The new-task update is $\Delta W_t := A_t B_t$, where $B_t\in\mathbb{R}^{r\times d_{\mathrm{out}}}$ is fixed and
the expectation is taken over the random initialization of $A_t\in\mathbb{R}^{d_{\mathrm{in}}\times r}$.
The goal is to bound the expected cosine similarity between $g_s$ and $\Delta W_t$.

Using $\langle U, V \rangle_F = \mathrm{Tr}(U^\top V)$ and the cyclic property, we have
\begin{equation}\label{eq:p-compact}
\big|\langle g_s, \Delta W_t \rangle_F \big|
=
\big|\langle A_t^\top g_s, B_t \rangle_F \big|
\le
\|A_t^\top g_s\|_F \, \|B_t\|_F .
\end{equation}
Normalizing by Frobenius norms yields
\begin{equation}\label{eq:p-ratio}
\frac{\langle g_s, \Delta W_t \rangle_F}
{\|g_s\|_F \, \|\Delta W_t\|_F}
\le
\frac{\|A_t^\top g_s\|_F}{\|g_s\|_F}
\cdot
\frac{\|B_t\|_F}{\|A_t B_t\|_F}.
\end{equation}

\noul{Numerator Bound.} 
Using the moment identity $\mathbb{E}[A_t A_t^\top] = \sigma^2 r I_{d_{\mathrm{in}}}$ from \eqref{eq:moments_formal}:
\begin{align}
\mathbb{E}_{A_t}\!\left[\|A_t^\top g_s\|_F^2\right]
&=
\mathbb{E}_{A_t}\!\left[\mathrm{Tr}\!\left(g_s^\top A_t A_t^\top g_s\right)\right]
=
\mathrm{Tr}\!\left(g_s^\top \mathbb{E}[A_t A_t^\top] g_s\right)
\nonumber\\
&=
\mathrm{Tr}\!\left(g_s^\top (\sigma^2 r I_{d_{\mathrm{in}}}) g_s\right)
=
\sigma^2 r \,\|g_s\|_F^2 .
\label{eq:p-first2}
\end{align}
By Jensen’s inequality, the expected norm is bounded by
\begin{equation}\label{eq:p-first1}
\mathbb{E}_{A_t}\!\left[\frac{\|A_t^\top g_s\|_F}{\|g_s\|_F}\right]
\le
\sqrt{\mathbb{E}_{A_t}\!\left[\frac{\|A_t^\top g_s\|_F^2}{\|g_s\|_F^2}\right]}
= \sigma \sqrt{r}.
\end{equation}

\noul{Denominator Bound.} 
Applying the concentration result from \eqref{eq:concentration_formal}, for a given $\epsilon \in (0,1)$, there exists an event $\mathcal{E}$ with probability at least $1-\delta$ such that:
\begin{equation}\label{eq:p-den-lb}
\|A_t B_t\|_F^2
\ge
(1-\epsilon)\, \sigma^2\, d_{\mathrm{in}}\, \|B_t\|_F^2.
\end{equation}
On this event, the second term of \eqref{eq:p-ratio} satisfies
\begin{equation}\label{eq:p-second}
\frac{\|B_t\|_F}{\|A_t B_t\|_F}
\le
\frac{1}{\sigma\sqrt{(1-\epsilon)\, d_{\mathrm{in}}}} .
\end{equation}

\noul{Expected Interference.} 
Since the cosine similarity is bounded by $1$ via Cauchy-Schwarz, we split the expectation over $\mathcal{E}$ and its complement $\mathcal{E}^c$.
Combining \eqref{eq:p-first1} and \eqref{eq:p-second}:
\begin{align}
\mathbb{E}_{A_t}\!\left[ \frac{\langle g_s, \Delta W_t \rangle_F}{\|g_s\|_F , \|\Delta W_t\|_F} \right]
&\le 
\frac{\sigma \sqrt{r}}{\sigma \sqrt{(1-\epsilon) d_{\mathrm{in}}}} + \mathbb{P}(\mathcal{E}^c) 
\\&= \frac{1}{\sqrt{1-\epsilon}} \sqrt{\frac{r}{d_{\mathrm{in}}}} + \delta \nonumber
=O\!\left(\sqrt{\tfrac{r}{d_{\mathrm{in}}}}\right).
\end{align}

\subsection{Proofs of Proposition 2} 

Let $A_s, A_t \in \mathbb{R}^{d_{\mathrm{in}}\times r}$ be independent random matrices for distinct tasks $s \neq t$.
Let $\widetilde{B}_s:=B_s\odot M_s$ and $\widetilde{B}_t:=B_t\odot M_t$. 
The task-specific updates are defined as $\Delta_s := A_s \widetilde{B}_s$ and $\Delta_t := A_t \widetilde{B}_t$.
The goal is to prove the normalized inner product of independent $\Delta_s$ and $\Delta_t$ converges to $0$ in probability.

The Frobenius inner product is given by: \begin{equation} \langle \Delta_s, \Delta_t \rangle_F = \mathrm{Tr}(\Delta_s^\top \Delta_t) = \mathrm{Tr}(\widetilde{B}_s^\top A_s^\top A_t \widetilde{B}_t). \end{equation}
Let $X := A_s^\top A_t \in \mathbb{R}^{r \times r}$ represent the interaction matrix between the two random projections. By the sub-multiplicative property of the Frobenius norm,
\begin{equation}\label{eq:p2-ratio}
\frac{|\langle \Delta_s,\Delta_t\rangle_F|}{\|\Delta_s\|_F\|\Delta_t\|_F}
\le
\|X\|_F\cdot
\frac{\|\widetilde{B}_s\|_F}{\|A_s\widetilde{B}_s\|_F}\cdot
\frac{\|\widetilde{B}_t\|_F}{\|A_t\widetilde{B}_t\|_F}.
\end{equation}

\noul{Interaction Matrix.}
For each $(k,\ell)$, $X_{k\ell} = \sum_{i=1}^{d_{\mathrm{in}}}(A_s)_{ik}(A_t)_{i\ell}$ is a sum of $d_{\mathrm{in}}$ independent random variables with mean $0$, and $\mathbb{E}[X_{k\ell}^2] = d_{\mathrm{in}} \sigma^4$.
Thus, the expected squared Frobenius norm is 
$\mathbb{E}\|X\|_F^2 = \sum_{k=1}^r \sum_{\ell=1}^r \mathbb{E}[X_{k\ell}^2] = r^2 d_{\mathrm{in}} \sigma^4$.
By Markov’s inequality applied to $\|X\|_F^2$, it follows that
\begin{equation}\label{eq:p2-X}\|X\|_F = O_p\!\left(r \sigma^2 \sqrt{d_{\mathrm{in}}}\right).\end{equation}

\noul{Update Norms.}
Using the concentration result from \eqref{eq:concentration_formal}, for any $\epsilon \in (0,1)$, the following lower bounds hold with high probability as $d_{\mathrm{in}} \to \infty$:
\begin{equation}\label{eq:p2-den}
\|\Delta_s\|_F^2 \ge (1-\epsilon) \sigma^2 d_{\mathrm{in}} \|\widetilde{B}_s\|_F^2, \quad \|\Delta_t\|_F^2 \ge (1-\epsilon) \sigma^2 d_{\mathrm{in}} \|\widetilde{B}_t\|_F^2.
\end{equation}
This implies that the norm ratios are bounded by 
\begin{equation}\label{eq:p2-ratio-den} \frac{\|\widetilde{B}_s\|_F}{\|\Delta_s\|_F} \le \frac{1}{\sigma \sqrt{(1-\epsilon)d_{\mathrm{in}}}}, \quad \frac{\|\widetilde{B}_t\|_F}{\|\Delta_t\|_F} \le \frac{1}{\sigma \sqrt{(1-\epsilon)d_{\mathrm{in}}}}. \end{equation}

\noul{Orthogonality.}
Plugging \eqref{eq:p2-X} and \eqref{eq:p2-ratio-den} into \eqref{eq:p2-ratio} yields 
\begin{equation}\label{eq:p2-final}
\begin{split}
\frac{|\langle \Delta_s,\Delta_t\rangle_F|}
{\|\Delta_s\|_F\|\Delta_t\|_F}
&\le
\frac{O_p(r \sigma^2 \sqrt{d_{\mathrm{in}}})}{(1-\epsilon)\sigma^2d_{\mathrm{in}}} \\
&=
O_p\!\left(\frac{r}{\sqrt{d_{\mathrm{in}}}}\right)
\xrightarrow[]{\mathrm{p}} 0,
\end{split}
\end{equation}
Since $r = o(\sqrt{d_{\mathrm{in}}})$ (which is satisfied if $r$ is constant or growing slower than the square root of the input dimension), the cosine similarity converges in probability to zero.

\subsection{\rcolor{Local Model-Level Interpretation}}
\label{appendix:model_level}

\rcolor{
Let $f_\theta(x)$ denote the logit vector of the model and let $\Delta\theta_t$ be the update introduced at round $t$. For small updates, a first-order expansion gives
\begin{equation}
f_{\theta+\Delta\theta_t}(x)-f_\theta(x)
=
J_\theta(x)\Delta\theta_t + R_2(x),
\end{equation}
where $J_\theta(x)=\nabla_\theta f_\theta(x)$ is the parameter Jacobian, and $R_2(x)$ is the second-order Taylor remainder, which captures the higher-order nonlinear effect of the update. 
If $f_\theta$ is locally $L$-smooth around $\theta$, then
\begin{equation}
\|R_2(x)\|_2 \le \frac{L}{2}\|\Delta\theta_t\|_2^2.
\end{equation}
Therefore,
\begin{equation}
\|f_{\theta+\Delta\theta_t}(x)-f_\theta(x)\|_2
\le
\|J_\theta(x)\|_2\|\Delta\theta_t\|_2
+
\frac{L}{2}\|\Delta\theta_t\|_2^2.
\end{equation}
Taking expectation over old data $x\sim \mathcal{D}_{\le t-1}$ gives
\begin{equation}
\mathbb{E}
\|f_{\theta+\Delta\theta_t}(x)-f_\theta(x)\|_2
\le
G_{\mathrm{old}}\|\Delta\theta_t\|_2
+
\frac{L}{2}\|\Delta\theta_t\|_2^2,
\end{equation}
where $G_{\mathrm{old}}=\mathbb{E}\|J_\theta(x)\|_2$.
For \sys{}, $\Delta\theta_t$ is controlled by the masked coefficient energy $E_t=\|M_t\odot B_t\|_F$. 
Limiting $E_t$ through low-rank structure and sparse masking bounds the Jacobian-projected output drift on old
distributions, providing a model-level interpretation of controlled forgetting under local smoothness and small-update assumptions. 
}

\section{Implementation Details} 

\subsection{\sys{} Implementation} \label{appendix:our_implementation}
Low-rank modules are applied to all linear layers in the MLP-based malware detection model.
For the transformer-based CodeT5 used in binary analysis, those modules are inserted into key subcomponents, including the self-attention projections (\texttt{q}, \texttt{k}, \texttt{v}, \texttt{o}) and the feed-forward layers (\texttt{wi}, \texttt{wo}).
We apply a real-valued learnable mask $M_t$ to modulate the new-task LoRA matrices, \rcolor{where rank $r$ is set to $512$ in both applications}. 
The mask is initialized with a sparsity ratio: $0.01$ in malware detection for strict forgetting control, and $1.0$ in binary analysis for a more progressive accumulation of new knowledge.
\rcolor{The initialization facilitates convergence, while the effective sparsity is learned automatically.}
There are several hyperparameters involved in performing the knowledge arbitration.
For malware detection, the sparsity-balancing parameter in~\autoref{equ:sparsity} is set to $\lambda = 1.0$, and the supervised loss coefficient and group-lasso regularization coefficient in~\autoref{equ:overall_loss} to $\eta = 1.0$ and $\mu = 2\times10^{-6}$, respectively.
For binary analysis, we use the same supervised loss coefficient $\eta = 1.0$ but adjust $\lambda = 0.1$ and $\mu = 1.0$ to accommodate the higher capacity of the transformer-based model.


\subsection{CL Baselines} \label{appendix:cl_baseline} 
For distillation-based baselines, the model is fine-tuned on each new task by augmenting the standard cross-entropy objective with a knowledge-distillation term, formulated as $\mathcal{L}=\lambda_{\mathrm{ce}}\,\mathcal{L}_{\mathrm{ce}}+\lambda_{\mathrm{kd}}\,\mathcal{L}_{\mathrm{kd}}$.
Different methods define $\mathcal{L}_{\mathrm{kd}}$ in distinct ways: LwF uses softened output logits from the teacher, PODNet constrains feature representations across layers, and Co\textsuperscript{2}L aligns pairwise embedding similarities. 

\textbf{LwF:} 
The distillation is calculated as the cross-entropy between the softened teacher and student probabilities, where are 
$p_t^{(i)} = \frac{\exp(z_t^{(i)}/T)}{\sum_j \exp(z_t^{(j)}/T)}$ and 
$p_s^{(i)} = \frac{\exp(z_s^{(i)}/T)}{\sum_j \exp(z_s^{(j)}/T)}$.
The temperature $T>1$ smooths the predictive distribution and conveys information about inter-class relationships. 
Through experiments over $T \in \{1.5, 2.0, 4.0\}$, we find that smaller temperatures produce overly peaked teacher distributions with limited
relational transfer, whereas larger temperatures over-flatten the distribution and reduce the effectiveness of distillation; accordingly, we use $T=2.0$.
Based on experimental observations, we determine distillation weights that provide a balance between different training objectives.
We select $\lambda_{\mathrm{ce}}=1.0$ and
$\lambda_{\mathrm{kd}}=1.0$ for malware detection, and $\lambda_{\mathrm{ce}}=1.0$ and $\lambda_{\mathrm{kd}}=0.1$ for binary analysis, where the distillation loss is averaged over all non-\texttt{[PAD]} tokens for sequence-generation tasks.

\textbf{PODNet:} 
Its distillation term matches $\ell_2$-normalized, pooled activations between the student and teacher across multiple layers and pooling scales.
We follow this distillation formulation to implement PODNet but omit its local similarity classifier designed for class-incremental learning.
For malware detection, we align the normalized activations from two hidden layers via chunk-wise pooling. 
For binary analysis, the spatial pyramid is achieved with a lightweight token-mean alignment module applied to the encoder’s last four layers, and feature matching is performed through mean-squared alignment of teacher and student embeddings. 
The PODNet loss weight is examined in the range
$\lambda_{\mathrm{pod}} \in \{0.5, 0.1, 0.2\}$ to control the strength of representation matching.
In practice, we fix $\lambda_{\mathrm{pod}}=0.1$, which provides sufficient regularization without over-constraining new-task adaptation.

\textbf{Co\textsuperscript{2}L:} 
It combines a supervised contrastive learning component with an instance-wise relation distillation module:
$\lambda_{\mathrm{con}}\mathcal{L}_{\mathrm{supcon}}
+\lambda_{\mathrm{ird}}\mathcal{L}_{\mathrm{IRD}}$.
While $\mathcal{L}_{\mathrm{supcon}}$ shapes the representation space by pulling together positive pairs and separating negatives, $\mathcal{L}_{\mathrm{IRD}}$ specifically preserves the pairwise similarity patterns between student and teacher embeddings. 
To balance supervised learning, contrastive alignment, and instance-wise relation distillation, we examine loss weights in $\{0.5, 1.0, 2.0\}$.
For malware detection, we use the penultimate-layer representations as embedding vectors for contrastive alignment. 
The loss weights are set to $\lambda_{\mathrm{ce}}{=}1.0$, $\lambda_{\mathrm{con}}{=}1.0$, and $\lambda_{\mathrm{ird}}{=}0.5$.   
In the binary analysis experiments, 
the encoder representations are mean-pooled across tokens and passed through a two-layer projection head for contrastive alignment.
The overall objective is weighted by $\lambda_{\mathrm{ce}}{=}0.5$, $\lambda_{\mathrm{con}}{=}1.0$, and $\lambda_{\mathrm{ird}}{=}0.5$, respectively. 

For architecture-based methods, \textbf{ResAdapt} introduces lightweight plug-in residual adapters into the backbone network while preserving the pretrained weights.  
Each adapter augments the representational capacity of the frozen layers through residual bottleneck projections:
$\mathrm{Adapter}(x) = x + U\!\big(\mathrm{ReLU}(D(\mathrm{LN}(x)))\big)$, where $\mathrm{LN}$ denotes layer normalization, $D$ and $U$ are down- and up-projection matrices defining the bottleneck.
To avoid relying on domain labels, we continuously fine-tune the separate adapter. 
For tuning, we vary its bottleneck dimension $b \in \{64, 128, 256, 512\}$ to balance the new and old threats, especially for competitive new-task performance compared to other baselines.
For malware detection, the adapters are injected into intermediate layers of the MLP backbone, with the bottleneck dimension set to $64$. 
For binary analysis, the residual adapter is extended to the transformer-based CodeT5 architecture, inserting adapters after each \textsc{T5Block} in both the encoder and decoder stacks and using a bottleneck dimension of $512$.


\subsection{Aggregation Baselines} \label{appendix:aggregate_baseline}

The two ensemble baselines: 
(1) \textbf{Ensemble Voting} computes the final prediction by taking the majority label among all models’ predictions, i.e.,  $\hat{y} = \mathrm{mode}\!\big(\{\hat{y}_t\}_{t=1}^n\big)$ where $\hat{y}_t$ is the prediction from the $t^{th}$ candidate model. 
(2) \textbf{Ensemble Averaging} averages logits from all models and applies a sigmoid function, expressed as $\hat{y} = \sigma\!\left(\tfrac{1}{n}\sum_{t=1}^n z_t(x)\right)$, 
where $z_t(x)$ denotes the logit output of the $t^{th}$ model.

The four model merging baselines: 
(1) \textbf{Weight Averaging} computes the element-wise weighted mean of the parameters of $n$ candidate models and can be formulated as 
$\boldsymbol{\theta}_{\mathrm{m}}=\tfrac{1}{n}\sum_{t=1}^{n}\boldsymbol{\theta}_{t}$, 
where $\boldsymbol{\theta}_{t}$ denotes the parameter vector of the $t$-th model and $\boldsymbol{\theta}_{\mathrm{m}}$ the merged parameters.
(2) \textbf{Task Arithmetic} constructs task vectors of the $n$ models relative to a pretrained base model and adds their scaled sum back to the base parameters, given by 
$\boldsymbol{\theta}_{\mathrm{m}}=\boldsymbol{\theta}_{\mathrm{base}}+\alpha\sum_{t=1}^{n}\boldsymbol{\tau}_{t}$, where $\boldsymbol{\tau}_{t}=\boldsymbol{\theta}_{t}-\boldsymbol{\theta}_{\mathrm{base}}$ is the task vector of model $t$, $\alpha$ is a scaling coefficient, and $\boldsymbol{\theta}_{\mathrm{base}}$ is the parameter vector of the base model.
we set $\alpha{=}0.3$.
(3) \textbf{Ties-Merging} extends Task Arithmetic by pruning parameter entries with conflicting signs and retaining only the top-$K\%$ magnitude components before combination. 
For malware detection, we set $K{=}5$ and $\alpha{=}0.3$; for binary analysis, we use $K{=}50$ and $\alpha{=}0.8$.
(4) \textbf{AdaMerging} has four variants, among which we adopt the most effective \emph{Layer-wise AdaMerging++} in its paper~\cite{yang2023adamerging}.
The method combines $K$ task models with the base model, formulated as 
$\boldsymbol{\theta}_{\mathrm{m}}^{\,l}=\boldsymbol{\theta}_{\mathrm{base}}^{\,l}+\sum_{k=1}^{K}\lambda_{k}^{\,l}\,\mathbf{T}_{k}^{\,l}$, 
where $\lambda_{k}^{\,l}$ is the learnable coefficient for task $k$ at layer $l$, 
and $\mathbf{T}_{k}^{\,l}=\boldsymbol{\theta}_{k}^{\,l}-\boldsymbol{\theta}_{\mathrm{base}}^{\,l}$ denotes the corresponding task vector.
The layer-specific coefficients are learned via entropy minimization on unlabeled samples, where entropy measures the uncertainty of a model’s predictive distribution, with higher values indicating less confidence.

\end{document}